\documentclass{article}
\pdfoutput=1

\usepackage{my_style}

\usepackage[utf8]{inputenc} % allow utf-8 input
\usepackage[T1]{fontenc}    % use 8-bit T1 fonts
\usepackage{times}

\usepackage[round]{natbib}
\usepackage{graphicx}
\usepackage{rotating}
\usepackage{array}
\usepackage{algorithm}
\usepackage{algorithmic}
\usepackage[labelfont=bf]{caption}

\usepackage{amsmath}
\usepackage{amssymb}
\usepackage{theorem}
\newcommand{\BlackBox}{\rule{1.5ex}{1.5ex}}  % end of proof
\newenvironment{proof}{\par\noindent{\bf Proof\ }}{\hfill\BlackBox\\[2mm]}

\newtheorem{theorem}{Theorem}

\newtheorem{lemma}[theorem]{Lemma}

\newcommand{\cbr}[1]{\left\{#1\right\}}

\renewcommand{\algorithmicrequire}{\textbf{Input:}}
\renewcommand{\algorithmicensure}{\textbf{Output:}}

\newcommand{\E}{\mathbb{E}}
\newcommand{\KL}{\text{KL}}
\newcommand{\JSD}{\text{JSD}}

\newcommand{\loss}{\mathcal{L}}

\newcolumntype{L}[1]{>{\raggedright\arraybackslash}p{#1}}
\newcolumntype{R}[1]{>{\raggedleft\arraybackslash}p{#1}}
\newcolumntype{C}[1]{>{\centering\let\newline\\\arraybackslash\hspace{0pt}}m{#1}}
\newcolumntype{?}{!{\vrule width 1pt}}
\makeatletter
\newcommand{\thickhline}{%
    \noalign {\ifnum 0=`}\fi \hrule height 1pt
    \futurelet \reserved@a \@xhline
}
\newcolumntype{"}{@{\hskip\tabcolsep\vrule width 1pt\hskip\tabcolsep}}
\makeatother
\newcommand{\intset}[1]{\cbr{1..n}}

\usepackage[colorlinks,pdfpagelabels,plainpages=false]{hyperref}
\usepackage{rotating}

\usepackage{xcolor}
\definecolor{dark-red}{rgb}{0.4,0.15,0.15}
\definecolor{dark-blue}{rgb}{0.15,0.15,0.4}
\hypersetup{
   colorlinks, linkcolor={dark-blue},
   citecolor={dark-blue}, urlcolor={medium-blue}
}

\usepackage{scrextend}
\usepackage[inline]{enumitem}
\usepackage{booktabs}
\usepackage{bm}
\usepackage{amsmath}
\usepackage{amssymb}
\usepackage{amsfonts}
\usepackage{mathtools}
\usepackage{comment}
\usepackage{subfigure}
\usepackage{url}

\newcommand{\mbf}[1]{{\boldsymbol{\mathbf{#1}}}}
\renewcommand{\bm}{\mbf}

\usepackage{parskip}

\usepackage{multirow}
\usepackage{titlefoot}
\usepackage{gensymb}

\title{\bf On Unifying Deep Generative Models}

\renewcommand{\footnotesize}{\normalsize}

\author{
  Zhiting Hu$^{1,2}$,~~ Zichao Yang$^{1}$,~~ Ruslan Salakhutdinov$^{1}$,~~ Eric Xing$^{2}$
  \and
  \texttt{\small\{zhitingh,zichaoy,rsalakhu\}@cs.cmu.edu, eric.xing@petuum.com}
  \and 
  {\small Carnegie Mellon University$^{1}$,~~ Petuum Inc.$^{2}$}
}

\begin{document}
\date{}

\maketitle

\begin{abstract}

\begin{sloppypar}
Deep generative models have achieved impressive success in recent years. 
Generative Adversarial Networks (GANs) and Variational Autoencoders (VAEs), as emerging families for generative model learning, have largely been considered as two distinct paradigms and received extensive independent studies respectively. 
This paper aims to establish formal connections between GANs and VAEs through a new formulation of them. 
We interpret sample generation in GANs as performing posterior inference, and show that GANs and VAEs involve minimizing KL divergences of respective posterior and inference distributions with opposite directions, extending the two learning phases of classic wake-sleep algorithm, respectively. The unified view provides a powerful tool to analyze a diverse set of existing model variants, and enables to transfer techniques across research lines in a principled way. For example, we apply the importance weighting method in VAE literatures for improved GAN learning, and enhance VAEs with an adversarial mechanism that leverages generated samples. Experiments show generality and effectiveness of the transfered techniques. 
\end{sloppypar}

\end{abstract}

\section{Introduction}
Deep generative models define distributions over a set of variables organized in multiple layers. Early forms of such models dated back to works on hierarchical Bayesian models~\citep{neal1992connectionist} and neural network models such as Helmholtz machines~\citep{dayan1995helmholtz}, originally studied in the context of unsupervised learning, latent space modeling, etc. Such models are usually trained via an EM style framework, using either a variational inference~\citep{jordan1999introduction} or a data augmentation~\citep{tanner1987calculation} algorithm. Of particular relevance to this paper is the classic wake-sleep algorithm dates by~\citet{hinton1995wake} for training Helmholtz machines, as it explored an idea of minimizing a pair of KL divergences in opposite directions of the posterior and its approximation.

In recent years there has been a resurgence of interests in deep generative modeling. The emerging approaches, including Variational Autoencoders (VAEs)~\citep{kingma2013auto}, Generative Adversarial Networks (GANs)~\citep{goodfellow2014generative}, Generative Moment Matching Networks (GMMNs)~\citep{li2015generative,dziugaite2015training}, auto-regressive neural networks~\citep{larochelle2011neural,oord2016pixel}, and so forth, have led to impressive results in a myriad of applications, such as image and text generation~\citep{radford2015unsupervised,hu2017controllable,van2016conditional}, disentangled representation learning~\citep{chen2016infogan,kulkarni2015deep}, and semi-supervised learning~\citep{salimans2016improved,kingma2014semi}. 

The deep generative model literature has largely viewed these approaches as distinct model training paradigms. For instance, GANs aim to achieve an equilibrium between a generator and a discriminator; while VAEs are devoted to maximizing a variational lower bound of the data log-likelihood. A rich array of theoretical analyses and model extensions have been developed independently for GANs~\citep{arjovsky2017towards,arora2017generalization,salimans2016improved,nowozin2016f} and VAEs~\citep{burda2015importance,chen2017variational,hu2017controllable}, respectively. A few works attempt to combine the two objectives in a single model for improved inference and sample generation~\citep{mescheder2017adversarial,larsen2015autoencoding,makhzani2015adversarial,sonderby2016amortised}. Despite the significant progress specific to each method, it remains unclear how these apparently divergent approaches connect to each other in a principled way.

In this paper, we present a new formulation of GANs and VAEs that connects them under a unified view, 
and links them back to the classic wake-sleep algorithm. We show that GANs and VAEs involve minimizing opposite KL divergences of respective posterior and inference distributions, and extending the sleep and wake phases, respectively, for generative model learning. More specifically, we develop a reformulation of GANs that interprets {\it generation} of samples as performing posterior {\it inference}, leading to an objective that resembles variational inference as in VAEs. As a counterpart, VAEs in our interpretation contain a {\it degenerated} adversarial mechanism that blocks out generated samples and only allows real examples for model training. 
 
The proposed interpretation provides a useful tool to analyze the broad class of recent GAN- and VAE-based algorithms, enabling perhaps a more principled and unified view of the landscape of generative modeling. For instance, one can easily extend our formulation to subsume InfoGAN~\citep{chen2016infogan} that additionally infers hidden representations of examples, VAE/GAN joint models~\citep{larsen2015autoencoding,che2017mode} that offer improved generation and reduced mode missing, and adversarial domain adaptation (ADA)~\citep{ganin2016domain,purushotham2017variational} that is traditionally framed in the discriminative setting. 

The close parallelisms between GANs and VAEs further ease transferring techniques that were originally developed for improving each individual class of models, to in turn benefit the other class. We provide two examples in such spirit: 1) Drawn inspiration from importance weighted VAE~(IWAE)~\citep{burda2015importance}, we straightforwardly derive importance weighted GAN~(IWGAN) that maximizes a tighter lower bound on the marginal likelihood compared to the vanilla GAN. 2) Motivated by the GAN adversarial game we activate the originally degenerated discriminator in VAEs, resulting in a full-fledged model that adaptively leverages both real and fake examples for learning. Empirical results show that the techniques imported from the other class are generally applicable to the base model and its variants, yielding consistently better performance.

\section{Related Work}\label{sec:related}
There has been a surge of research interest in deep generative models in recent years, with remarkable progress made in understanding several class of algorithms. The wake-sleep algorithm~\citep{hinton1995wake} is one of the earliest general approaches for learning deep generative models. The algorithm incorporates a separate inference model for posterior approximation, and aims at maximizing a variational lower bound of the data log-likelihood, or equivalently, minimizing the KL divergence of the approximate posterior and true posterior. However, besides the wake phase that minimizes the KL divergence w.r.t the generative model, the sleep phase is introduced for tractability that minimizes instead the {\it reversed} KL divergence w.r.t the inference model. Recent approaches such as NVIL~\citep{mnih2014neural} and VAEs~\citep{kingma2013auto} are developed to maximize the variational lower bound w.r.t both the generative and inference models jointly. 
To reduce the variance of stochastic gradient estimates, VAEs leverage reparametrized gradients. Many works have been done along the line of improving VAEs. \citet{burda2015importance} develop importance weighted VAEs to obtain a tighter lower bound.  
As VAEs do not involve a sleep phase-like procedure, the model cannot leverage samples from the generative model for model training. \citet{hu2017controllable} combine VAEs with an extended sleep procedure that exploits generated samples for learning.

Another emerging family of deep generative models is the Generative Adversarial Networks (GANs) \citep{goodfellow2014generative}, in which a discriminator is trained to distinguish between real and generated samples and the generator to confuse the discriminator. The adversarial approach can be alternatively motivated in the perspectives of approximate Bayesian computation~\citep{gutmann2014statistical} and density ratio estimation~\citep{mohamed2016learning}. The original objective of the generator is to minimize the log probability of the discriminator correctly recognizing a generated sample as fake. This is equivalent to {\it minimizing a lower bound} on the Jensen-Shannon divergence (JSD) of the generator and data distributions~\citep{goodfellow2014generative,nowozin2016f,huszar2016infogan,li2016gans}. Besides, the objective suffers from vanishing gradient with strong discriminator. Thus in practice people have used another objective which maximizes the log probability of the discriminator recognizing a generated sample as real~\citep{goodfellow2014generative,arjovsky2017towards}. The second objective has the same optimal solution as with the original one. We base our analysis of GANs on the second objective as it is widely used in practice yet few theoretic analysis has been done on it. Numerous extensions of GANs have been developed, including combination with VAEs for improved generation~\citep{larsen2015autoencoding,makhzani2015adversarial,che2017mode}, and generalization of the objectives to minimize other $f$-divergence criteria beyond JSD~\citep{nowozin2016f,sonderby2016amortised}. The adversarial principle has gone beyond the generation setting and been applied to other contexts such as domain adaptation~\citep{ganin2016domain,purushotham2017variational}, and Bayesian inference~\citep{mescheder2017adversarial,tran2017deep,huszar2017variational,rosca2017variational} which uses implicit variational distributions in VAEs and leverage the adversarial approach for optimization. This paper starts from the basic models of GANs and VAEs, and develops a general formulation that reveals underlying connections of different classes of approaches including many of the above variants, yielding a unified view of the broad set of deep generative modeling. 

This paper considerably extends the conference version~\citep{hu2018unifying} by generalizing the unified framework to a broader set of GAN- and VAE-variants, providing a more complete and consistent view of the various models and algorithms, adding more discussion of the symmetric view of generation and inference, and re-organizing thbe presentation to make the theory development clearer.

\section{Bridging the Gap}
The structures of GANs and VAEs are at the first glance quite different from each other. VAEs are based on the variational inference approach, and include an explicit inference model that reverses the generative process defined by the generative model. On the contrary, in traditional view GANs lack an inference model, but instead have a discriminator that judges generated samples. In this paper, a key idea to bridge the gap is to interpret the generation of samples in GANs as {\it performing inference}, and the discrimination as a generative process that produces real/fake labels. The resulting new formulation reveals the connections of GANs to traditional variational inference. The reversed generation-inference interpretations between GANs and VAEs also expose their correspondence to the two learning phases in the classic wake-sleep algorithm. 

For ease of presentation and to establish a systematic notation for the paper, we start with a new interpretation of {\it Adversarial Domain Adaptation} (ADA)~\citep{ganin2016domain}, the application of adversarial approach in the domain adaptation context. We then show GANs are a special case of ADA, followed with a series of analysis linking GANs, VAEs, and their variants in our formulation.

\subsection{Adversarial Domain Adaptation (ADA)}
Given two domains, one source domain with labeled data and one target domain without labels, ADA aims to transfer prediction knowledge learned from the source domain to the target domain, by learning domain-invariant features~\citep{ganin2016domain,qin2017adversarial,purushotham2017variational}. That is, it learns a feature extractor whose output cannot be distinguished by a discriminator between the source and target domains. 

\begin{figure} 
%\vspace{-15pt}
  \centering 
  \includegraphics[width=0.8\linewidth]{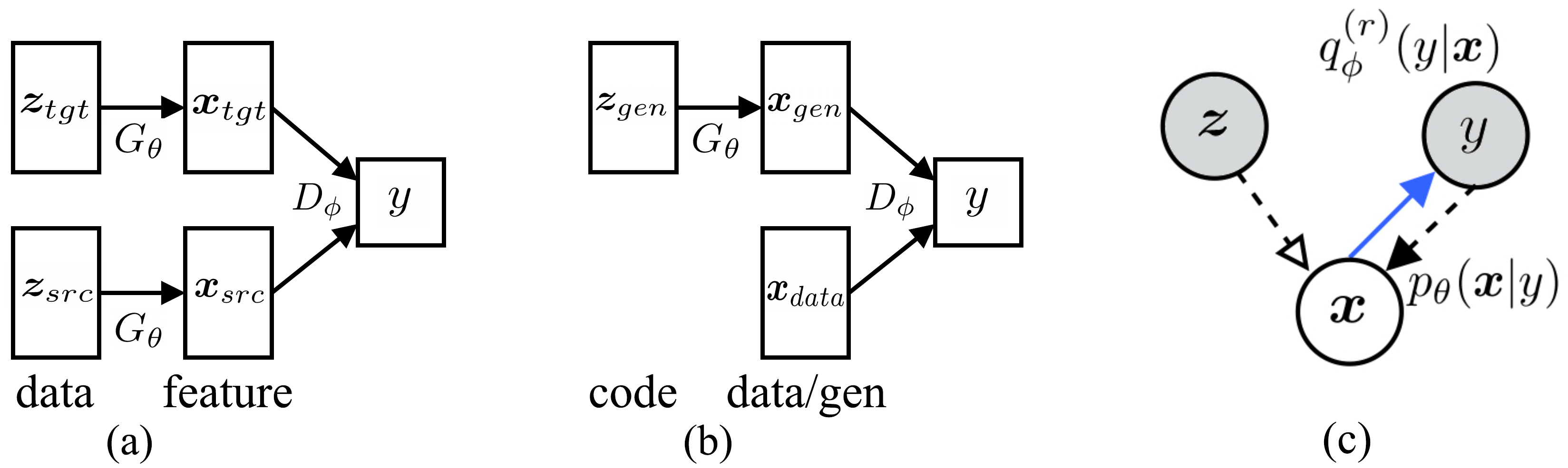}
  %\vspace{-20pt}
  \caption{{\bf (a)} Conventional view of ADA. To make direct correspondence to GANs, we use $\bm{z}$ to denote the data and $\bm{x}$ the feature. Subscripts {\it src} and {\it tgt} denote the source and target domains, respectively. {\bf (b)} Conventional view of GANs. The code space of the real data domain is degenerated. {\bf (c)} Schematic graphical model of both ADA and GANs (Eqs.\ref{eq:ada-obj-phi}-\ref{eq:ada-obj-theta}). Arrows with solid lines denote generative process; arrows with dashed lines denote inference; hollow arrows denote deterministic transformation leading to implicit distributions; and blue arrows denote adversarial mechanism that involves respective conditional distribution $q$ and its reverse $q^{r}$, e.g., $q(y|\bm{x})$ and $q^{r}(y|\bm{x})$ (denoted as $q^{(r)}(y|\bm{x})$ for short). Note that in GANs we have interpreted $\bm{x}$ as a latent variable and $(\bm{z},y)$ as visible.} 
\label{fig:gm_ada_gan}
%\vspace{-8pt}
\end{figure}

We first review the conventional formulation of ADA. 
Figure~\ref{fig:gm_ada_gan}(a) illustrates the computation flow. Let $\bm{z}$ be a data example either in the source or target domain, and $y\in\{0,1\}$ the domain indicator with $y=0$ indicating the target domain and $y=1$ the source domain. The domain-specific data distributions can then be denoted as a conditional distribution $p(\bm{z}|y)$. The feature extractor $G_\theta$ parameterized with $\bm{\theta}$ maps data $\bm{z}$ to feature $\bm{x}=G_{\theta}(\bm{z})$. To enforce domain invariance of feature $\bm{x}$, a discriminator $D_{\phi}$ is learned. 
Specifically, $D_{\phi}(\bm{x})$ outputs the probability that $\bm{x}$ comes from the source domain. The discriminator is trained to maximize the binary classification accuracy of recognizing the domains:
\begin{equation}
\small
\begin{split}
\max\nolimits_{\bm{\phi}}\loss_{\phi} &= \E_{\bm{x}=G_{\theta}(\bm{z}), \bm{z}\sim p(\bm{z}|y=1)}\left[ \log D_\phi(\bm{x}) \right] + \E_{\bm{x} = G_{\theta}(\bm{z}), \bm{z}\sim p(\bm{z}|y=0)}\left[ \log (1-D_\phi(\bm{x}))  \right].
\end{split}
\label{eq:obj-ada-d-expand}
\end{equation}
The feature extractor $G_\theta$ is then trained to fool the discriminator:
\begin{equation}
\small
\begin{split}
\max\nolimits_{\bm{\theta}}\loss_{\theta} &= \E_{\bm{x}=G_\theta(\bm{z}), \bm{z}\sim p(\bm{z}|y=1)}\left[ \log (1-D_\phi(\bm{x})) \right] + \E_{\bm{x} = G_{\theta}(\bm{z}), \bm{z}\sim p(\bm{z}|y=0)}\left[ \log D_\phi(\bm{x})  \right].
\end{split}
\label{eq:obj-ada-g-expand}
\end{equation}
We omit the additional loss on $\bm{\theta}$ that improves the accuracy of the original classification problem based on source-domain features~\citep{ganin2016domain}.

With the background of conventional formulation, we now frame our new interpretation of ADA. The data distribution $p(\bm{z}|y)$ and deterministic transformation $G_\theta$ together form an {\it implicit} distribution over $\bm{x}$, denoted as $p_{\theta}(\bm{x}|y)$:
\begin{equation}
\begin{split}
\bm{x} \sim p_{\theta}(\bm{x}|y)  ~~\Leftrightarrow~~
\bm{x} = G_\theta(\bm{z}),~~ \bm{z}\sim p(\bm{z}|y).
\end{split}
\label{eq:p-implicit}
\end{equation}
The distribution $p_\theta(\bm{x}|y)$ is intractable for evaluating likelihood but easy to sample from. Let $p(y)$ be the distribution of the domain indicator $y$, e.g., a uniform distribution as in Eqs.\eqref{eq:obj-ada-d-expand}-\eqref{eq:obj-ada-g-expand}. The discriminator defines a conditional distribution $q_\phi(y|\bm{x})=D_\phi(\bm{x})$. Let $q_\phi^{r}(y|\bm{x}) = q_\phi(1-y|\bm{x})$ be the reversed distribution over domains. The objectives of ADA can then be rewritten as (omitting the constant scale factor $2$):
%\begin{equation}
%\begin{split}
\begin{align}
\small
\label{eq:ada-obj-phi}
&\max\nolimits_{\bm{\phi}} \loss_{\phi} = \E_{p_\theta(\bm{x}|y)p(y)}\left[ \log q_\phi(y|\bm{x})  \right] \\
&\max\nolimits_{\bm{\theta}} \loss_{\theta} = \E_{p_\theta(\bm{x}|y) p(y)}\left[ \log q_\phi^{r}(y|\bm{x}) \right].
%\end{split}
\label{eq:ada-obj-theta}
\end{align}
%\end{equation}
Note that $\bm{z}$ is encapsulated in the implicit distribution $p_\theta(\bm{x}|y)$ (Eq.\ref{eq:p-implicit}). 
The only difference of the objectives for $\bm{\theta}$ from $\bm{\phi}$ is the replacement of $q(y|\bm{x})$ with $q^r(y|\bm{x})$. This is where the adversarial mechanism comes about. We defer deeper interpretation of the new objectives in the next subsection.

\subsection{Generative Adversarial Networks (GANs)}
GANs~\citep{goodfellow2014generative} can be seen as a special case of ADA. Taking image generation for example, intuitively, we want to transfer the properties of real image (source domain) to generated image (target domain), making them indistinguishable to the discriminator. Figure~\ref{fig:gm_ada_gan}(b) shows the conventional view of GANs. 

Formally, 
$\bm{x}$ now denotes a real example or a generated sample, $\bm{z}$ is the respective latent code. For the generated sample domain ($y=0$), the implicit distribution $p_\theta(\bm{x}|y=0)$ is defined by the prior of $\bm{z}$ and the generator $G_\theta(\bm{z})$ (Eq.\ref{eq:p-implicit}), which is also denoted as $p_{g_\theta}(\bm{x})$ in the literature. For the real example domain ($y=1$), the code space and generator are {\it degenerated},
and we are directly presented with a fixed distribution $p(\bm{x}|y=1)$, which is just the real data distribution $p_{data}(\bm{x})$. Note that $p_{data}(\bm{x})$ is also an implicit distribution and allows efficient empirical sampling. In summary, the conditional distribution over $\bm{x}$ is constructed as
\begin{equation}
\small
\begin{split}
p_\theta(\bm{x}|y) = 
\begin{cases}
p_{g_\theta}(\bm{x}) & y=0 \\
p_{data}(\bm{x})  &  y=1.
\end{cases}
\end{split}
\label{eq:gan-px_y}
\end{equation}
Here, free parameters $\bm{\theta}$ are only associated with $p_{g_\theta}(\bm{x})$ of the generated sample domain, while $p_{data}(\bm{x})$ is constant. As in ADA, discriminator $D_\phi$ is simultaneously trained to infer the probability that $\bm{x}$ comes from the real data domain. That is, $q_\phi(y=1|\bm{x})=D_\phi(\bm{x})$. 

With the established correspondence between GANs and ADA, we can see that the objectives of GANs are precisely expressed as Eqs.\eqref{eq:ada-obj-phi}-\eqref{eq:ada-obj-theta}. 
To make this clearer, we recover the classical form by unfolding over $y$ and plugging in conventional notations. For instance, the objective of the generative parameters $\bm{\theta}$ in Eq.\eqref{eq:ada-obj-theta} is translated into
\begin{equation}
\small
\begin{split}
\max\nolimits_{\bm{\theta}} \loss_{\theta} &= \E_{p_\theta(\bm{x}|y=0) p(y=0)}\left[ \log q_\phi^{r}(y=0|\bm{x}) \right] + \E_{p_\theta(\bm{x}|y=1) p(y=1)}\left[ \log q_\phi^{r}(y=1|\bm{x}) \right] \\
%&= \frac{1}{2} \E_{\bm{x}=G_\theta(\bm{z}),\bm{z}\sim p(\bm{z}|y=0)}\left[ \log D_\phi(\bm{x}) \right] + \frac{1}{2} \E_{\bm{x}\sim p_{data}(\bm{x})}\left[ \log (1-D_\phi(\bm{x})) \right] \\
&= \frac{1}{2} \E_{\bm{x}=G_\theta(\bm{z}),\bm{z}\sim p(\bm{z}|y=0)}\left[ \log D_\phi(\bm{x}) \right] + const,
\end{split}
\label{eq:gan-obj}
\end{equation}
where $p(y)$ is uniform and results in the constant scale factor $1/2$. As noted in section~\ref{sec:related}, we focus on the unsaturated objective for the generator~\citep{goodfellow2014generative}, as it is commonly used in practice yet still lacks systematic analysis.

\subsubsection{New Interpretation of GANs} 
Let us take a closer look into the form of Eqs.\eqref{eq:ada-obj-phi}-\eqref{eq:ada-obj-theta}. If we treat $y$ as a visible variable while $\bm{x}$ as latent (as in ADA), Eq.\eqref{eq:ada-obj-phi} closely resembles the M-step in a \emph{variational EM}~\citep{bernardo2003variational} learning procedure. That is, we are ``reconstructing'' the real/fake indicator $y$ with the ``generative distribution'' $q_\phi(y|\bm{x})$, conditioning on $\bm{x}$ inferred from the ``variational distribution'' $p_\theta(\bm{x}|y)$. Similarly, Eq.\eqref{eq:ada-obj-theta} is in analogue with the E-step with the ``generative distribution'' now being $q_\phi^r(y|\bm{x})$, except that the KL divergence regularization between the ``variational distribution'' $p_\theta(\bm{x}|y)$ and some ``prior'' $p_{prior}(\bm{x})$ over $\bm{x}$, i.e.,  $\KL(p_\theta(\bm{x}|y)\|p_{prior}(\bm{x}))$ is missing. We take this view and reveal the connections between GANs and variational learning further in the following.

\paragraph{Schematic graphical model representation}
Before going a step further, we first illustrate such generative and inference processes in GANs in Figure~\ref{fig:gm_ada_gan}(c). We have introduced new visual elements to augment the conventional graphical model representation, for example, hollow arrows for expressing implicit distributions, and blue arrows for adversarial objectives.  We found such a graphical representation can precisely express various deep generative models in our new perspective, and make the connections between them clearer. We will see more such graphical representations later.

\vspace{4pt}

We continue to analyze the objective for $\bm{\theta}$ (Eq.\ref{eq:ada-obj-theta}).
Let $\bm{\Theta}_0 = (\bm{\theta}=\bm{\theta}_0, \bm{\phi}=\bm{\phi}_0)$ be the state of the parameters from the last iteration. At current iteration, a natural idea is to treat the marginal distribution over $\bm{x}$ at $\bm{\Theta}_0$ as the ``prior'':
\begin{equation}
\small
\begin{split}
p_{\theta_0}(\bm{x}) := \E_{p(y)}[p_{\theta_0} (\bm{x}|y)].
\end{split}
\end{equation}
As above, $q_{\phi_0}^r(y|\bm{x})$ in Eq.\eqref{eq:ada-obj-theta} can be interpreted as the ``likelihood'' function in variational EM. We then can construct the ``posterior'':
\begin{equation}
\small
\begin{split}
q^{r}(\bm{x}|y) \propto q_{\phi_0}^{r}(y|\bm{x})p_{\theta_0}(\bm{x}).
\end{split}
\end{equation}
We have the following results in terms of the gradient w.r.t $\bm{\theta}$:
\begin{lemma}
Let $p(y)$ be the uniform distribution. 
The update of $\bm{\theta}$ at $\bm{\theta}_0$ has
\begin{equation}
\small
\begin{split}
&\nabla_{\theta} \Big[ - \E_{p_{\theta}(\bm{x}|y)p(y)}\left[ \log q^{r}_{\phi_0}(y|\bm{x}) \right] \Big] \Big|_{\bm{\theta}=\bm{\theta}_0}\Big. = \\ 
&\nabla_{\theta} \Big[ \E_{p(y)}\left[ \KL\left( p_{\theta}(\bm{x}|y) \big\| q^{r}(\bm{x}|y)\right) \right] - \JSD\left( p_\theta(\bm{x}|y=0) \big\| p_\theta(\bm{x}|y=1) \right) \Big] \Big|_{\bm{\theta}=\bm{\theta}_0}\Big. ,
\end{split}
\label{eq:thm-1-result}
\end{equation}
where $\KL(\cdot\|\cdot)$ and $\JSD(\cdot\|\cdot)$ are the Kullback-Leibler and Jensen-Shannon Divergences, respectively.
\end{lemma}
Proofs are provided in the supplements (section~\ref{supp:sec:lemma-1}). The result offers new insights into the GAN generative model learning:

\begin{figure} 
%\vspace{-8pt}
  \centering 
  \includegraphics[width=0.95\linewidth]{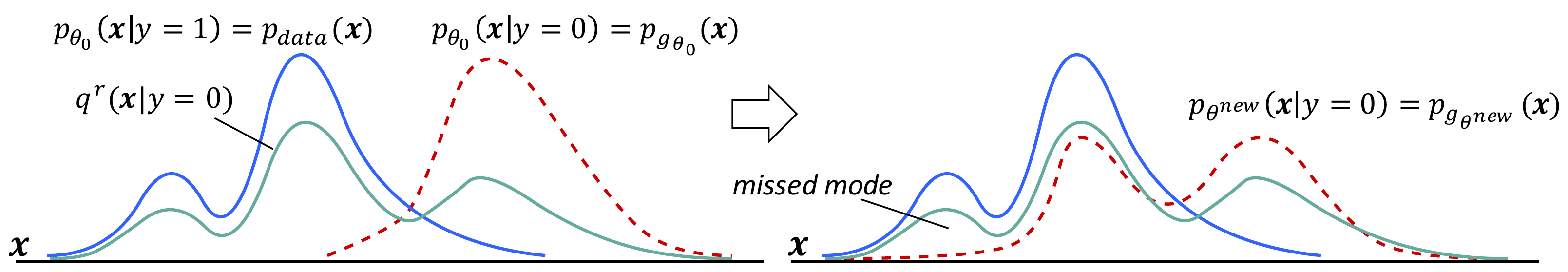}
  %\vspace{-10pt}
  \caption{\small One optimization step of the parameter $\bm{\theta}$ through Eq.\eqref{eq:thm-1-result} at point $\bm{\theta}_0$. The posterior $q^{r}(\bm{x}|y)$ is a mixture of $p_{\theta_0}(\bm{x}|y=0)$ (blue) and $p_{\theta_0}(\bm{x}|y=1)$ (red in the left panel) with the mixing weights induced from $q^r_{\phi_0}(y|\bm{x})$. Minimizing the KLD drives $p_\theta(\bm{x}|y=0)$ towards the respective mixture $q^{r}(\bm{x}|y=0)$ (green), resulting in a new state where $p_{\theta^{new}}(\bm{x}|y=0)=p_{g_{\theta^{new}}}(\bm{x})$ (red in the right panel) gets closer to $p_{\theta_0}(\bm{x}|y=1)=p_{data}(\bm{x})$. Due to the asymmetry of KLD, $p_{g_{\theta^{new}}}(\bm{x})$ missed the smaller mode of the mixture $q^{r}(\bm{x}|y=0)$ which is a mode of $p_{data}(\bm{x})$.}
  \label{fig:lemma-interpret}
  %\vspace{-6pt}
\end{figure}

\begin{addmargin}[-10pt]{0pt}
%[-25pt]{0pt} % indent -20pt left, 0pt right
  \begin{itemize}
  \item {\bf Resemblance to variational inference. }As discussed above, we see $\bm{x}$ as the latent variable and $p_\theta(\bm{x}|y)$ the variational distribution that approximates the true ``posterior'' $q^{r}(\bm{x}|y)$.
Therefore, optimizing the generator $G_{\theta}$ is equivalent to minimizing the KL divergence between the inference distribution and the posterior (a standard from of variational inference), minus a JSD between the distributions $p_{g_\theta}(\bm{x})$ and $p_{data}(\bm{x})$. The interpretation also helps to reveal the connections between GANs and VAEs, as discussed later.
  \item {\bf The JSD term. } The negative JSD term is due to the introduction of the prior $p_{\theta_0}(\bm{x})$. This term pushes $p_{g_\theta}(\bm{x})$ away from $p_{data}(\bm{x})$, which acts oppositely from the KLD term. However, we show in the supplementary that the JSD term is upper bounded by the KL term (section~\ref{supp:sec:jsd-bound}). Thus, if the KL term is sufficiently minimized, the magnitude of the JSD also decreases. Note that we do not mean the JSD is insignificant or negligible. Instead any conclusions drawn from Eq.\eqref{eq:thm-1-result} should take the JSD term into account.
  \item {\bf Training dynamics. } The component with $y=1$ in the KL divergence term is
  \begin{equation}
  \small
\begin{split}
  \KL\left( p_{\theta}(\bm{x}|y=1) \| q^{r}(\bm{x}|y=1)\right)=\KL\left( p_{data}(\bm{x}) \| q^{r}(\bm{x}|y=1)\right)
  \end{split}
  \end{equation}
which is a constant. The active component associated with the parameter $\bm{\theta}$ to optimize is with $y=0$, i.e., 
  \begin{equation}
  \small
\begin{split}
  \KL\left( p_{\theta}(\bm{x}|y=0) \| q^{r}(\bm{x}|y=0)\right)=\KL\left( p_{g_\theta}(\bm{x}) \| q^{r}(\bm{x}|y=0)\right).
  \end{split}
  \end{equation}
On the other hand, by definition, $p_{\theta_0}(\bm{x})=(p_{g_{\theta_0}}(\bm{x})+p_{data}(\bm{x}))/2$ is a mixture of $p_{g_{\theta_0}}(\bm{x})$ and $p_{data}(\bm{x})$, and thus the posterior $q^{r}(\bm{x}|y=0) \propto q_{\phi_0}^{r}(y=0|\bm{x})p_{\theta_0}(\bm{x})$ is also a mixture of $p_{g_{\theta_0}}(\bm{x})$ and $p_{data}(\bm{x})$ with mixing weights induced from the discriminator $q^{r}_{\phi_0}(y=0|\bm{x})$.  Thus, minimizing the KL divergence in effect drives $p_{g_\theta}(\bm{x})$ to a mixture of $p_{g_{\theta_0}}(\bm{x})$ and $p_{data}(\bm{x})$. Since $p_{data}(\bm{x})$ is fixed, $p_{g_\theta}(\bm{x})$ gets closer to $p_{data}(\bm{x})$. Figure~\ref{fig:lemma-interpret} illustrates the training dynamics schematically.
  \item {\bf Explanation of missing mode issue. } JSD is a symmetric divergence measure while KL is asymmetric. The missing mode behavior widely observed in GANs~\citep{metz2017unrolled,che2017mode} is thus explained by the asymmetry of the KL which tends to concentrate $p_{\theta}(\bm{x}|y)$ to large modes of $q^{r}(\bm{x}|y)$ and ignore smaller ones. See Figure~\ref{fig:lemma-interpret} for the illustration. Concentration to few large modes also facilitates GANs to generate sharp and realistic samples.
  \item {\bf Optimality assumption of the discriminator. } Previous theoretical works have typically assumed (near) optimal discriminator~\citep{goodfellow2014generative,arjovsky2017towards}:
  \begin{equation}
  \small
  \begin{split}
  q_{\phi_0}(y|\bm{x}) \approx \frac{p_{\theta_0}(\bm{x}|y=1)}{p_{\theta_0}(\bm{x}|y=0)+p_{\theta_0}(\bm{x}|y=1)} = \frac{p_{data}(\bm{x})}{p_{g_{\theta_0}}(\bm{x})+p_{data}(\bm{x})},
  \end{split}
  \label{eq:opt-disc}
  \end{equation}
  which can be unwarranted in practice due to limited expressiveness of the discriminator~\citep{arora2017generalization}. In contrast, our result does not rely on the optimality assumptions. Indeed, our result is a generalization of the previous theorem in~\citep{arjovsky2017towards}, which is recovered by plugging Eq.\eqref{eq:opt-disc} into Eq.\eqref{eq:thm-1-result}:
  \begin{equation}
   \small
  \begin{split}
  \nabla_{\theta} \Big[ - \E_{p_{\theta}(\bm{x}|y)p(y)}\left[ \log q^{r}_{\phi_0}(y|\bm{x}) \right] \Big] \Big|_{\bm{\theta}=\bm{\theta}_0}\Big. =  \nabla_{\theta} \left[ \frac{1}{2}\KL\left( p_{g_\theta}\| p_{data}\right) - \JSD\left( p_{g_\theta} \| p_{data} \right) \right] \Big|_{\bm{\theta}=\bm{\theta}_0}\Big., 
  \end{split}
  \label{eq:lemma-1-optimal}
  \end{equation}
  which gives simplified explanations of the training dynamics and the missing mode issue, but only when the discriminator meets certain optimality criteria. Our generalized result enables understanding of broader situations. For instance, when the discriminator distribution $q_{\phi_0}(y|\bm{x})$ gives uniform guesses, or when $p_{g_\theta}=p_{data}$ that is indistinguishable by the discriminator, the gradients of the KL and JSD terms in Eq.\eqref{eq:thm-1-result} cancel out, which stops the generator learning. 
  \end{itemize}
\end{addmargin}

\begin{figure} 
%\vspace{-15pt}
  \centering 
  \includegraphics[width=0.8\linewidth]{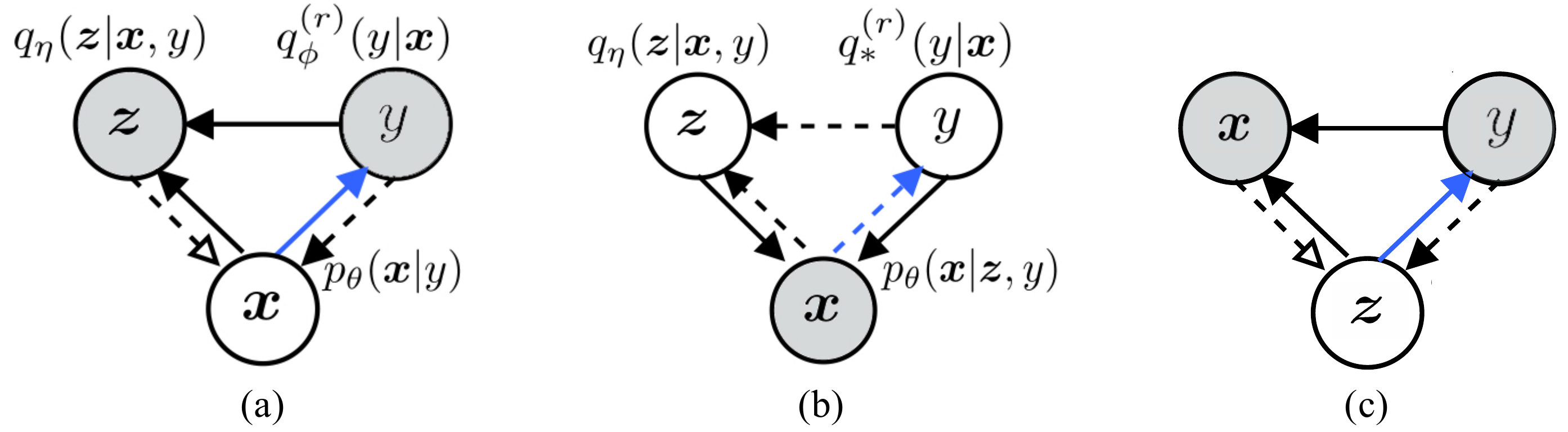}
  %\vspace{-20pt}
  \caption{{\bf (a) }Schematic graphical model of InfoGAN (Eq.\ref{eq:infogan-obj}), which, compared to GANs, adds conditional generative process of code $\bm{z}$ with distribution $q_\eta(\bm{z}|\bm{x},y)$. {\bf (b)} VAEs (Eq.\ref{eq:vae-new}), which is obtained by swapping the generative and inference processes of InfoGAN, i.e., in terms of the schematic graphical model, swapping solid-line arrows (generative process) and dashed-line arrows (inference) of (a). {\bf (c) }Adversarial Autoencoder (AAE), which is obtained by swapping data $\bm{x}$ and code $\bm{z}$ in InfoGAN. } 
\label{fig:gm_infogan_aae_vae}
%\vspace{-8pt}
\end{figure}

\subsubsection{InfoGAN}
\citet{chen2016infogan} developed InfoGAN that additionally recovers the code $\bm{z}$ given sample $\bm{x}$. This can straightforwardly be formulated in our framework by introducing an extra conditional $q_{\eta}(\bm{z}|\bm{x}, y)$ parameterized by $\bm{\eta}$. As discussed above, GANs assume a degenerated code space for real examples, thus $q_{\eta}(\bm{z}|\bm{x}, y=1)$ is defined to be fixed without free parameters to learn, and $\bm{\eta}$ is only associated to the $y=0$ case. Further, as in Figure~\ref{fig:gm_ada_gan}(c), $\bm{z}$ is treated as a visible variable. Thus $q_{\eta}(\bm{z}|\bm{x}, y)$ augments the generative process, leading to a full likelihood $q_{\eta}(\bm{z}|\bm{x}, y)q_\phi(y|\bm{x})$. The InfoGAN is then recovered by extending Eqs.\eqref{eq:ada-obj-phi}-\eqref{eq:ada-obj-theta} as follows:
\begin{equation}
\small
\begin{split}
&\max\nolimits_{\bm{\phi}} \loss_{\phi} = \E_{p_\theta(\bm{x}|y)p(y)}\left[ \log q_{\eta}(\bm{z}|\bm{x}, y) q_\phi(y|\bm{x})  \right] \\
&\max\nolimits_{\bm{\theta},\bm{\eta}} \loss_{\theta,\eta} = \E_{p_\theta(\bm{x}|y) p(y)}\left[ \log q_{\eta}(\bm{z}|\bm{x}, y) q_\phi^{r}(y|\bm{x}) \right].
\end{split}
\label{eq:infogan-obj}
\end{equation}
Again, note that $\bm{z}$ is encapsulated in the implicit distribution $p_\theta(\bm{x}|y)$. The model is expressed as the schematic graphical model in Figure~\ref{fig:gm_ada_gan}(d). 

The resulting $\bm{z}$-augmented posterior is $q^{r}(\bm{x}|\bm{z},y)\propto q_{\eta_0}(\bm{z}|\bm{x},y)q^{r}_{\phi_0}(y|\bm{x})p_{\theta_0}(\bm{x})$. The result in the form of Lemma.1 still holds:
\begin{equation}
\small
\begin{split}
&\nabla_{\theta} \Big[ - \E_{p_{\theta}(\bm{x}|y)p(y)}\left[ \log q_{\eta_0}(\bm{z}|\bm{x},y) q^{r}_{\phi_0}(y|\bm{x}) \right] \Big] \Big|_{\bm{\theta}=\bm{\theta}_0}\Big. = \\ 
& \nabla_{\theta} \Big[ \E_{p(y)}\left[ \KL\left( p_{\theta}(\bm{x}|y) \big\| q^{r}(\bm{x}|\bm{z},y) \right) \right] - \JSD\left( p_\theta(\bm{x}|y=0) \big\| p_\theta(\bm{x}|y=1) \right) \Big] \Big|_{\bm{\theta}=\bm{\theta}_0}\Big. ,
\end{split}
\label{eq:infogan-lemma}
\end{equation}

\subsubsection{Adversarial Autoencoder (AAE) and CycleGAN}\label{sec:aae}

The new formulation is also generally applicable to other GAN-related variants, such as Adversarial Autoencoder (AAE)~\citep{makhzani2015adversarial}, Predictability Minimization~\citep{schmidhuber1992learning}, and cycleGAN~\citep{zhu2017unpaired}. 

Specifically, AAE is recovered by simply swapping the code variable $\bm{z}$ and the data variable $\bm{x}$ of InfoGAN in the graphical model, as shown in Figure~\ref{fig:gm_infogan_aae_vae}(c). In other words, AAE is precisely an InfoGAN that treats the code $\bm{z}$ as a latent variable to be adversarially regularized, and the data/generation variable $\bm{x}$ as a visible variable. To make this clearer, in the supplementary we demonstrate how the schematic graphical model of Figure~\ref{fig:gm_infogan_aae_vae}(c) can directly be translated into the mathematical formula of AAE~\citep{makhzani2015adversarial}. Predictability Minimization (PM)~\citep{schmidhuber1992learning} resembles AAE and is also discussed in the supplementary materials.

InfoGAN and AAE thus are a symmetric pair that exchanges data and code spaces. Further, instead of considering $\bm{x}$ and $\bm{z}$ as data and code spaces respectively, if we use both $\bm{x}$ and $\bm{z}$ to model data spaces of two modalities, and combine the objectives of InfoGAN and AAE as a joint model, we recover the cycleGAN model~\citep{zhu2017unpaired} that performs transformation between the two modalities. In particular, the objectives of AAE (Eq.\ref{supp:eq:aae-obj} in the supplementary) are precisely the objectives that train the cycleGAN model to translate $\bm{x}$ into $\bm{z}$,
and the objectives of InfoGAN (Eq.\ref{eq:infogan-obj}) are used to train the reversed translation from $\bm{z}$ to $\bm{x}$.

\subsection{Variational Autoencoders (VAEs)}
We next explore the second family 
of deep generative modeling, namely, the VAEs~\citep{kingma2013auto}. The resemblance of GAN learning to variational inference (Lemma.1) suggests strong relations between VAEs and GANs. We build correspondence between them, and show that VAEs involve minimizing a KLD in an opposite direction to that of GANs, with a degenerated adversarial discriminator.

The conventional definition of VAEs is written as:
\begin{equation}
\small
\begin{split}
\max\nolimits_{\bm{\theta},\bm{\eta}} \mathcal{L}^{\text{vae}}_{\theta,\eta} = \E_{p_{data}(\bm{x})}\Big[ \E_{\tilde{q}_{\eta}(\bm{z}|\bm{x})} \left[ \log \tilde{p}_\theta(\bm{x}|\bm{z}) \right] - \KL(\tilde{q}_\eta(\bm{z}|\bm{x}) \| \tilde{p}(\bm{z})) \Big],
\end{split}
\label{eq:vae-ori}
\end{equation}
where $\tilde{p}_{\theta}(\bm{x}|\bm{z})$ is the generative model, $\tilde{q}_{\eta}(\bm{z}|\bm{x})$ the inference model, and $\tilde{p}(\bm{z})$ the prior. The parameters to learn are intentionally denoted with the notations of corresponding modules in GANs. VAEs appear to differ from GANs greatly as they use only real examples and lack adversarial mechanism. 

To connect to GANs, we assume a {\it perfect} discriminator $q_{*}(y|\bm{x})$ that always predicts $y=1$ with probability $1$ given real examples, and $y=0$ given generated samples. Again, for notational simplicity, let $q_{*}^{r}(y|\bm{x})=q_{*}(1-y|\bm{x})$ be the reversed distribution. Precisely as for GANs, in our formulation, the code space $\bm{z}$ for real examples are degenerated, and we are presented with the real data distribution $p_{data}(\bm{x})$ directly over $\bm{x}$. The composite conditional distribution of $\bm{x}$ is thus constructed as:
\begin{equation}
\small
\begin{split}
p_\theta(\bm{x}|\bm{z},y) = 
\begin{cases}
\tilde{p}_\theta(\bm{x}|\bm{z}) & y=0 \\
p_{data}(\bm{x})  &  y=1.
\end{cases}
\end{split}
\label{eq:vae-px_y}
\end{equation}
We can see the distribution differs slightly from its GAN counterpart $p_{\theta}(\bm{x}|y)$ in Eq.\eqref{eq:gan-px_y} and additionally accounts for the uncertainty of generating $\bm{x}$ given $\bm{z}$. In analogue to InfoGAN, we have the conditional over $\bm{z}$, namely, $q_\eta(\bm{z}|\bm{x},y)$, in which $q_\eta(\bm{z}|\bm{x},y=1)$ is constant due to the degenerated code space for $y=1$, and $q_\eta(\bm{z}|\bm{x},y=0)=\tilde{q}_{\eta}(\bm{z}|\bm{x})$ is associated with the free parameter $\bm{\eta}$. Finally, we extend the prior over $\bm{z}$ to define $p(\bm{z}|y)$ such that $p(\bm{z}|y=0)=\tilde{p}(\bm{z})$ and $p(\bm{z}|y=1)$ is again irrelevant.

We are now ready to reformulate the VAE objective in Eq.\eqref{eq:vae-ori}:
\begin{lemma}
Let $p_\theta(\bm{z},y|\bm{x}) \propto p_\theta(\bm{x}|\bm{z},y) p(\bm{z}|y)p(y)$. The VAE objective $\mathcal{L}^{\text{vae}}_{\theta,\eta}$ in Eq.\eqref{eq:vae-ori} is equivalent to (omitting the constant scale factor $2$):
\begin{equation}
\small
\begin{split}
\mathcal{L}^{\text{vae}}_{\theta,\eta} &=  \E_{p_{\theta_0}(\bm{x})}\Big[ \E_{\begin{subarray}{l} q_{\eta}(\bm{z}|\bm{x},y) q_{*}^{r}(y|\bm{x}) \end{subarray}} \left[ \log p_\theta(\bm{x}|\bm{z},y) \right] - \KL\left(q_\eta(\bm{z}|\bm{x},y) q_{*}^{r}(y|\bm{x}) \big\| p(\bm{z}|y)p(y)\right) \Big] \\
&= \E_{p_{\theta_0}(\bm{x})}\Big[ - \KL\left( q_{\eta}(\bm{z}|\bm{x},y)q_{*}^{r}(y|\bm{x}) \big\| p_\theta(\bm{z},y|\bm{x}) \right) \Big].
\end{split}
\label{eq:vae-new}
\end{equation}
\end{lemma}
We provide the proof in the supplementary materials (section~\ref{supp:sec:lemma-2}). 

Figure~\ref{fig:gm_infogan_aae_vae}(b) shows the schematic graphical model of the new interpretation of VAEs, where the only difference from InfoGAN is swapping the solid-line arrows (generative process) and dashed-line arrows (inference).
That is, InfoGAN and VAEs are \emph{a symmetric pair} in the sense of exchanging the generative and inference process.

Given a fake sample $\bm{x}$ from $p_{\theta_0}(\bm{x})$, the {\it reversed} perfect discriminator $q_{*}^{r}(y|\bm{x})$ always predicts $y=1$ with probability $1$, and the loss on fake samples is therefore degenerated to a constant (irrelevant to the free parameters). This blocks out fake samples from contributing to the model learning.

\begin{table}[t]
%\vspace{-15pt}
\centering
\small
\begin{tabular}{r l l l}
\cmidrule[\heavyrulewidth](lr){1-4}
  Components & ADA & GANs / InfoGAN & VAEs \\ \cmidrule(lr){1-4}
 $\bm{x}$ & features & data/generations  & data/generations \\ [0.07cm]
 $y$ & domain indicator & real/fake indicator & real/fake indicator (degenerated) \\ [0.07cm]
 $\bm{z}$ & data examples & code vector & code vector \\ \cmidrule(lr){1-4}
$p_{\theta}(\bm{x}|y)$ & feature distr. & {\bf [I]}\ \  generator, Eq.\ref{eq:gan-px_y}  &  {\bf [G]} $p_{\theta}(\bm{x}|\bm{z},y)$, generator, Eq.\ref{eq:vae-px_y} \\ [0.07cm]
$q_{\phi}(y|\bm{x})$ & discriminator & {\bf [G]} discriminator & {\bf [I]}\ \ \ $q_*(y|\bm{x})$, discriminator (degenerated) \\ [0.07cm]
$q_{\eta}(\bm{z}|\bm{x},y)$ &  --- & {\bf [G]} infer net (InfoGAN) & {\bf [I]}\ \ \ infer net \\ \cmidrule(lr){1-4}
%$p_{\theta_0}(\bm{x})$ (lemma 1) & same as GANs  & prior of $\bm{x}$ & prior of $\bm{x}$ \\
KLD to min & same as GANs & $\KL\left( p_{\theta}(\bm{x}|y) \| q^{r}(\bm{x}|y)\right)$ & $\KL\left( q_{\eta}(\bm{z}|\bm{x},y)q_{*}^{r}(y|\bm{x}) \| p_\theta(\bm{z},y|\bm{x}) \right)$ \\
\cmidrule[\heavyrulewidth](lr){1-4}
\end{tabular}
%\vspace{-10pt}
\caption{\small Correspondence between different approaches in the proposed formulation. The label ``[G]'' in bold indicates the respective component is involved in the generative process within our interpretation, while ``[I]'' indicates inference process. This is also expressed in the schematic graphical models in Figure~\ref{fig:gm_ada_gan}.}
\label{table:notations}
%\vspace{-10pt}
\end{table}

\subsection{Connecting the Two Families of GANs and VAEs}
Table~\ref{table:notations} summarizes the correspondence between the various methods.
Lemma.1 and Lemma.2 have revealed that both GANs and VAEs involve minimizing a KLD of respective inference and posterior distributions. Specifically, GANs involve minimizing the ${\text {\textit{KL}}}\left( p_{\theta}(\bm{x}|y) \big\| q^{r}(\bm{x}|y)\right) $ while VAEs the ${\text {\textit{KL}}}\left( q_{\eta}(\bm{z}|\bm{x},y)q_{*}^{r}(y|\bm{x}) \big\| p_\theta(\bm{z},y|\bm{x}) \right)$. This exposes new connections between the two model classes in multiple aspects, each of which in turn leads to a set of existing research or can inspire new research directions: 
\begin{addmargin}[-10pt]{0pt}
%[-25pt]{0pt} % indent -20pt left, 0pt right
\begin{enumerate}[label*=\arabic*)]
\item As discussed in Lemma.1, GANs now also relate to the variational inference algorithm as with VAEs, revealing a unified statistical view of the two classes. Moreover, the new perspective naturally enables many of the extensions of VAEs and vanilla variational inference algorithm to be transferred to GANs. We show an example in the next section.
\item The generator parameters $\bm{\theta}$ are placed in the opposite directions in the two KLDs. The asymmetry of KLD leads to distinct model behaviors. For instance, as discussed in Lemma.1, GANs are able to generate sharp images but tend to collapse to one or few modes of the data (i.e., mode missing). In contrast, the KLD of VAEs tends to drive generator to cover all modes of the data distribution but also small-density regions (i.e., mode covering), which tend to result in blurred samples. Such opposite behaviors naturally inspires combination of the two objectives to remedy the asymmetry of each of the KLDs, as discussed below.
\item VAEs within our formulation also include an adversarial mechanism as in GANs. The discriminator is perfect and degenerated, disabling generated samples to help with learning. This inspires activating the adversary to allow learning from samples. We present a simple possible way in the next section.
\item GANs and VAEs have inverted latent-visible treatments of $(\bm{z},y)$ and $\bm{x}$, since we interpret sample generation in GANs as posterior inference. Such inverted treatments strongly relates to the symmetry of the sleep and wake phases in the wake-sleep algorithm, as presented shortly. In section~\ref{sec:discussion}, we provide a more general discussion on a symmetric view of generation and inference.
\end{enumerate}
\end{addmargin}

\subsubsection{VAE/GAN Joint Models}
Previous work has explored combination of VAEs and GANs into joint models. As above, this can be naturally motivated by the opposite asymmetric behaviors of the KLDs that the two algorithms optimize respectively. Specifically, \citet{larsen2015autoencoding,pu2017symmetric} improve the sharpness of VAE generated images by adding the GAN objective that forces the generative model to focus on meaningful data modes. On the other hand, augmenting GANs with the VAE objective helps addressing the mode missing problem, which is studied in~\citep{che2017mode}.

\subsubsection{Implicit Variational Inference}\label{sec:implicit-inference}
Another recent line of research extends VAEs by using an implicit model as the variational distribution~\citep{mescheder2017adversarial,tran2017deep,huszar2017variational,rosca2017variational}. The idea naturally matches GANs under the unified view. In particular, in Eq.\eqref{eq:thm-1-result}, the ``variational distribution'' $p_\theta(\bm{x}|y)$ of GANs is also an implicit model. Such implicit variational distribution does not assume a particular distribution family (e.g., Gaussian distributions) and thus provides improved flexibility. Compared to GANs, the implicit variational inference in VAEs additionally forces the variational distribution to be close to a prior distribution. This is usually achieved by minimizing an adversarial loss between the two distribution, as in AAE (section~\ref{sec:aae}).

\subsection{Connecting to Wake Sleep Algorithm (WS)}
Wake-sleep algorithm~\citep{hinton1995wake} was proposed for learning deep generative models such as Helmholtz machines~\citep{dayan1995helmholtz}. WS consists of wake phase and sleep phase, which optimize the generative model and inference model, respectively. We follow the above notations, and introduce new notations $\bm{h}$ to denote general latent variables and $\bm{\lambda}$ to denote general parameters. The wake sleep algorithm is thus written as: 
\begin{equation}
\small
\begin{split}
&\text{Wake}:\quad \max\nolimits_{\bm{\theta}} \E_{q_{\lambda}(\bm{h}|\bm{x})p_{data}(\bm{x})}\left[ \log p_{\theta}(\bm{x}|\bm{h}) \right] \\ 
&\text{Sleep}:\quad \max\nolimits_{\bm{\lambda}} \E_{p_{\theta}(\bm{x}|\bm{h})p(\bm{h})}\left[ \log q_{\lambda}(\bm{h}|\bm{x}) \right].
\end{split}
\label{eq:wake-sleep}
\end{equation}
Briefly, the wake phase updates the generator parameters $\bm{\theta}$ by fitting $p_{\theta}(\bm{x}|\bm{h})$ to the real data and hidden code inferred by the inference model $q_{\lambda}(\bm{h}|\bm{x})$. On the other hand, the sleep phase updates the parameters $\bm{\lambda}$ based on the generated samples from the generator. \citet{hu2017controllable} have briefly discussed the relations between the WS, VAEs, and GANs algorithms. We formalize the discussion in the below.

The relations between WS and VAEs are clear in previous discussions~\citep{bornschein2014reweighted,kingma2013auto}. Indeed, WS was originally proposed to minimize the variational lower bound as in VAEs (Eq.\ref{eq:vae-ori}) with the sleep phase approximation~\citep{hinton1995wake}. Alternatively, VAEs can be seen as extending the wake phase. Specifically, if we let $\bm{h}$ be $\bm{z}$ and $\bm{\lambda}$ be $\bm{\eta}$, the wake phase objective recovers VAEs (Eq.\ref{eq:vae-ori}) in terms of generator optimization (i.e., optimizing $\bm{\theta}$). Therefore, we can see VAEs as generalizing the wake phase by also optimizing the inference model $q_\eta$, with additional prior regularization on code $\bm{z}$. 

On the other hand, GANs resemble the sleep phase. To make this clearer, let $\bm{h}$ be $y$ and $\bm{\lambda}$ be $\bm{\phi}$. This results in a sleep phase objective identical to that of optimizing the discriminator $q_{\phi}$ in Eq.\eqref{eq:ada-obj-phi}, which is to reconstruct $y$ given sample $\bm{x}$.
We thus can view GANs as generalizing the sleep phase by also optimizing the generative model $p_\theta$ to reconstruct reversed $y$. InfoGAN (Eq.\ref{eq:infogan-obj}) further extends to reconstruct the code $\bm{z}$. 

\subsection{The Relation Graph}
\begin{figure} 
%\vspace{-15pt}
  \centering 
  \includegraphics[width=\linewidth]{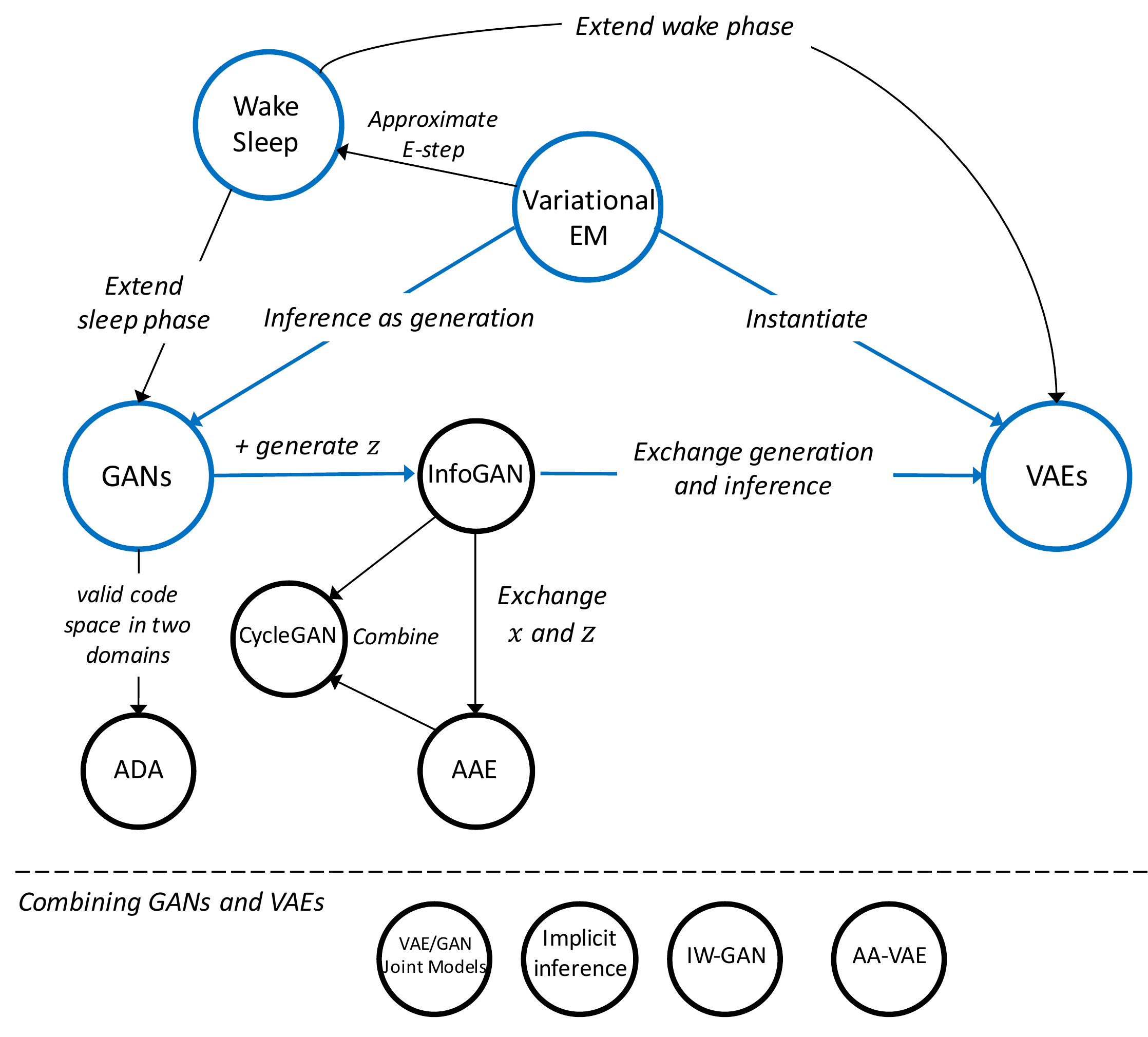}
  %\vspace{-20pt}
  \caption{Relations between deep generative models and algorithms discussed in the paper. The triangular relation in blue highlights the backbone of the unified framework. IW-GAN and AA-VAE are two extensions inspired from the connections between GANs and VAEs (section~\ref{sec:app}).} 
\label{fig:roadmap}
%\vspace{-8pt}
\end{figure}

We have presented the unified view that connects GANs and VAEs to classic variational EM and wake-sleep algorithms, and subsumes a broad set of their variants and extensions. Figure~\ref{fig:roadmap} depicts the essential relations between the various deep generative models and algorithms under our unified perspective. The generality of the proposed formulation offers a unified statistical insight of the broad landscape of deep generative modeling. 

One of the key ideas in our formulation is to treat the sample generation in GANs as performing posterior inference. Treating inference and generation as a symmetric pair leads to the triangular relation in blue in Figure~\ref{fig:roadmap}. We provide more discussion of the symmetric treatment in section~\ref{sec:discussion}.

\section{Applications: Transferring Techniques}\label{sec:app}
The above new formulations not only reveal the connections underlying the broad set of existing approaches, but also facilitate to exchange ideas and transfer techniques across the different families of models and algorithms. For instance, existing enhancements on VAEs can straightforwardly be applied to improve GANs, and vice versa. This section gives two examples. Here we only outline the main intuitions and resulting models, while providing the details in the supplement materials.

\subsection{Importance Weighted GANs (IWGAN)}
\citet{burda2015importance} proposed importance weighted autoencoder (IWAE) that maximizes a tighter lower bound on the marginal likelihood. Within our framework it is straightforward to develop importance weighted GANs by {\it copying} the derivations of IWAE side by side, with little adaptations. 
Specifically, the variational inference interpretation in Lemma.1 suggests GANs can be viewed as maximizing a lower bound of the marginal likelihood on $y$ (putting aside the negative JSD term):
\begin{equation}
\small
\begin{split}
\log q(y) &= \log \int p_{\theta}(\bm{x}|y)  \frac{q^{r}_{\phi_0}(y|\bm{x}) p_{\theta_0}(\bm{x})}{p_\theta(\bm{x}|y)} d\bm{x} \geq -\KL(p_{\theta}(\bm{x}|y) \| q^{r}(\bm{x}|y)) + const.
\end{split}
\label{eq:iwgan-lb}
\end{equation}
Following~\citep{burda2015importance}, we can derive a tighter lower bound through a $k$-sample importance weighting estimate of the marginal likelihood. 
With necessary approximations for tractability, optimizing the tighter lower bound results in the following update rule for the generator learning:
\begin{equation}
\small
\begin{split}
\nabla_\theta \mathcal{L}_k(y) = \E_{\bm{z}_1,\dots,\bm{z}_k \sim p(\bm{z}|y)} \left[ \sum\nolimits_{i=1}^{k} \widetilde{w_i} \nabla_\theta \log  q^r_{\phi_0}(y|\bm{x}(\bm{z}_i,\bm{\theta})) \right].
\end{split}
\label{eq:iwgan-update}
\end{equation}
As in GANs, only $y=0$ (i.e., generated samples) is effective for learning parameters $\bm{\theta}$. Compared to the vanilla GAN update (Eq.\eqref{eq:thm-1-result}), the only difference here is the additional importance weight $\widetilde{w_i}$ which is the normalization of $w_i = \frac{q^r_{\phi_0}(y|\bm{x}_i)}{q_{\phi_0}(y|\bm{x}_i)}$ over $k$ samples. Intuitively, the algorithm assigns higher weights to samples that are more realistic and fool the discriminator better, which is consistent to IWAE that emphasizes more on code states providing better reconstructions. \citet{hjelm2017boundary,che2017maximum} developed a similar sample weighting scheme for generator training, while their generator of discrete data depends on explicit conditional likelihood. In practice, the $k$ samples correspond to sample minibatch in standard GAN update. Thus the only computational cost added by the importance weighting method is by evaluating the weight for each sample, and is negligible. The discriminator is trained in the same way as in standard GANs.

\subsection{Adversary Activated VAEs (AAVAE)}
By Lemma.2, VAEs include a degenerated discriminator which blocks out generated samples from contributing to model learning. We enable adaptive incorporation of fake samples by activating the adversarial mechanism. 
Specifically, we replace the perfect discriminator $q_{*}(y|\bm{x})$ in VAEs with a discriminator network $q_{\phi}(y|\bm{x})$ parameterized with $\bm{\phi}$, resulting in an adapted objective of Eq.\eqref{eq:vae-new}:
\begin{equation}
\small
\begin{split}
\max_{\bm{\theta},\bm{\eta}} \mathcal{L}^{\text{aavae}}_{\theta,\eta} &=  \E_{p_{\theta_0}(\bm{x})}\left[ \E_{\begin{subarray}{l} q_{\eta}(\bm{z}|\bm{x},y) q_{\phi}^{r}(y|\bm{x}) \end{subarray}} \left[ \log p_\theta(\bm{x}|\bm{z},y) \right] - \KL(q_\eta(\bm{z}|\bm{x},y) q_{\phi}^{r}(y|\bm{x}) \| p(\bm{z}|y)p(y)) \right].
\end{split}
\label{eq:aavae-gen}
\end{equation}
As detailed in the supplementary material, the discriminator is trained in the same way as in GANs. 

The activated discriminator enables an effective data selection mechanism. First, AAVAE uses not only real examples, but also generated samples for training. Each sample is weighted by the inverted discriminator $q^{r}_{\phi}(y|\bm{x})$, so that only those samples that resemble real data and successfully fool the discriminator will be incorporated for training. This is consistent with the importance weighting strategy in IWGAN. Second, real examples are also weighted by $q^{r}_{\phi}(y|\bm{x})$. An example receiving large weight indicates it is easily recognized by the discriminator, which means the example is hard to be simulated from the generator. That is, AAVAE emphasizes more on harder examples. 

\begin{table}
%\vspace{-15pt}
\parbox{.31\linewidth}{
\centering
\small
\begin{tabular}{r L{1.2cm} l}
\cmidrule[\heavyrulewidth](lr){1-3}
%\hline
& GAN & IWGAN \\  \cmidrule(lr){1-3}
MNIST & 8.34$\pm$.03 & {\bf 8.45$\pm$.04} \\
SVHN & 5.18$\pm$.03 & {\bf 5.34$\pm$.03} \\
CIFAR10 & 7.86$\pm$.05 & {\bf 7.89$\pm$ .04} \\
\cmidrule[\heavyrulewidth](lr){1-3}
\end{tabular}
}
\hspace{0.2in} 
\parbox{.33\linewidth}{
\centering
\small
\begin{tabular}{R{1.1cm} L{1.5cm} l}
%\hline 
\cmidrule[\heavyrulewidth](lr){1-3}
& CGAN & IWCGAN \\  \cmidrule(lr){1-3}
MNIST &  0.985$\pm$.002 & {\bf 0.987$\pm$.002}  \\
SVHN & 0.797$\pm$.005 & {\bf 0.798$\pm$.006} \\
\cmidrule[\heavyrulewidth](lr){1-3}
\end{tabular}
}
\hspace{0.2in} 
\parbox{.25\linewidth}{
\centering
\small
\begin{tabular}{r L{0.9cm} l}
\cmidrule[\heavyrulewidth](lr){1-3}
&  SVAE & AASVAE \\  \cmidrule(lr){1-3}
1\% & 0.9412 & {\bf 0.9425} \\
10\% & 0.9768 & {\bf 0.9797} \\
\cmidrule[\heavyrulewidth](lr){1-3}
\end{tabular}
}
\vspace{-8pt}
\caption{\small {\bf Left}: Inception scores of GANs and the importance weighted extension. {\bf Middle}: Classification accuracy of the generations by conditional GANs and the IW extension. {\bf Right}: Classification accuracy of semi-supervised VAEs and the AA extension on MNIST test set, with $1\%$ and $10\%$ real labeled training data.}
\label{tab:exp}
\vspace{-5pt}
\end{table}
\begin{table}
\centering
\small
\begin{tabular}{r | l l | l l | l l}
\cmidrule[\heavyrulewidth](lr){1-7}
Train Data Size & VAE & AA-VAE & CVAE & AA-CVAE & SVAE & AA-SVAE\\  \cmidrule(lr){1-7}
1\% & -122.89 & {\bf -122.15} & -125.44 & {\bf -122.88} & -108.22 & {\bf -107.61} \\
10\% & -104.49 & {\bf -103.05} & -102.63 & {\bf -101.63} & -99.44 & {\bf -98.81} \\
100\% & -92.53 & {\bf -92.42} & -93.16 & {\bf -92.75} & --- & --- \\
\cmidrule[\heavyrulewidth](lr){1-7}
\end{tabular}
\vspace{-8pt}
\caption{\small Variational lower bounds on MNIST test set, trained on $1\%, 10\%$, and $100\%$ training data, respectively. In the semi-supervised VAE (SVAE) setting, remaining training data are used for unsupervised training.}
\label{tab:aa-exp}
%\vspace{-7pt}
\end{table}

\section{Experiments}
We conduct preliminary experiments to demonstrate the generality and effectiveness of the importance weighting (IW) and adversarial activating (AA) techniques. In this paper we do not aim at achieving state-of-the-art performance, but leave it for future work. In particular, we show the IW and AA extensions improve the standard GANs and VAEs, as well as several of their variants, respectively. We present the results here, and provide details of experimental setups in the supplements. 

\subsection{Importance Weighted GANs}
We extend both vanilla GANs and class-conditional GANs (CGAN) with the IW method.
The base GAN model is implemented with the DCGAN architecture and hyperparameter setting~\citep{radford2015unsupervised}. Hyperparameters are not tuned for the IW extensions. We use MNIST, SVHN, and CIFAR10 for evaluation. For vanilla GANs and its IW extension, we measure inception scores~\citep{salimans2016improved} on the generated samples.  
For CGANs we evaluate the accuracy of conditional generation~\citep{hu2017controllable} with a pre-trained classifier. Please see the supplements for more details.

Table~\ref{tab:exp}, left panel, shows the inception scores of GANs and IW-GAN, and the middle panel gives the classification accuracy of CGAN and and its IW extension. We report the averaged results $\pm$ one standard deviation over 5 runs. The IW strategy gives consistent improvements over the base models.

\subsection{Adversary Activated VAEs}
We apply the AA method on vanilla VAEs, class-conditional VAEs (CVAE), and semi-supervised VAEs (SVAE)~\citep{kingma2014semi}, respectively. We evaluate on the MNIST data.  
We measure the variational lower bound on the test set, with varying number of real training examples. For each batch of real examples, AA extended models generate equal number of fake samples for training. 

Table~\ref{tab:aa-exp} shows the results of activating the adversarial mechanism in VAEs. 
Generally, larger improvement is obtained with smaller set of real training data. Table~\ref{tab:exp}, right panel, shows the improved accuracy of AA-SVAE over the base semi-supervised VAE. 

\section{Discussions: Symmetric View of Generation and Inference}\label{sec:discussion}
\begin{figure} 
  \centering 
  \includegraphics[width=0.6\linewidth]{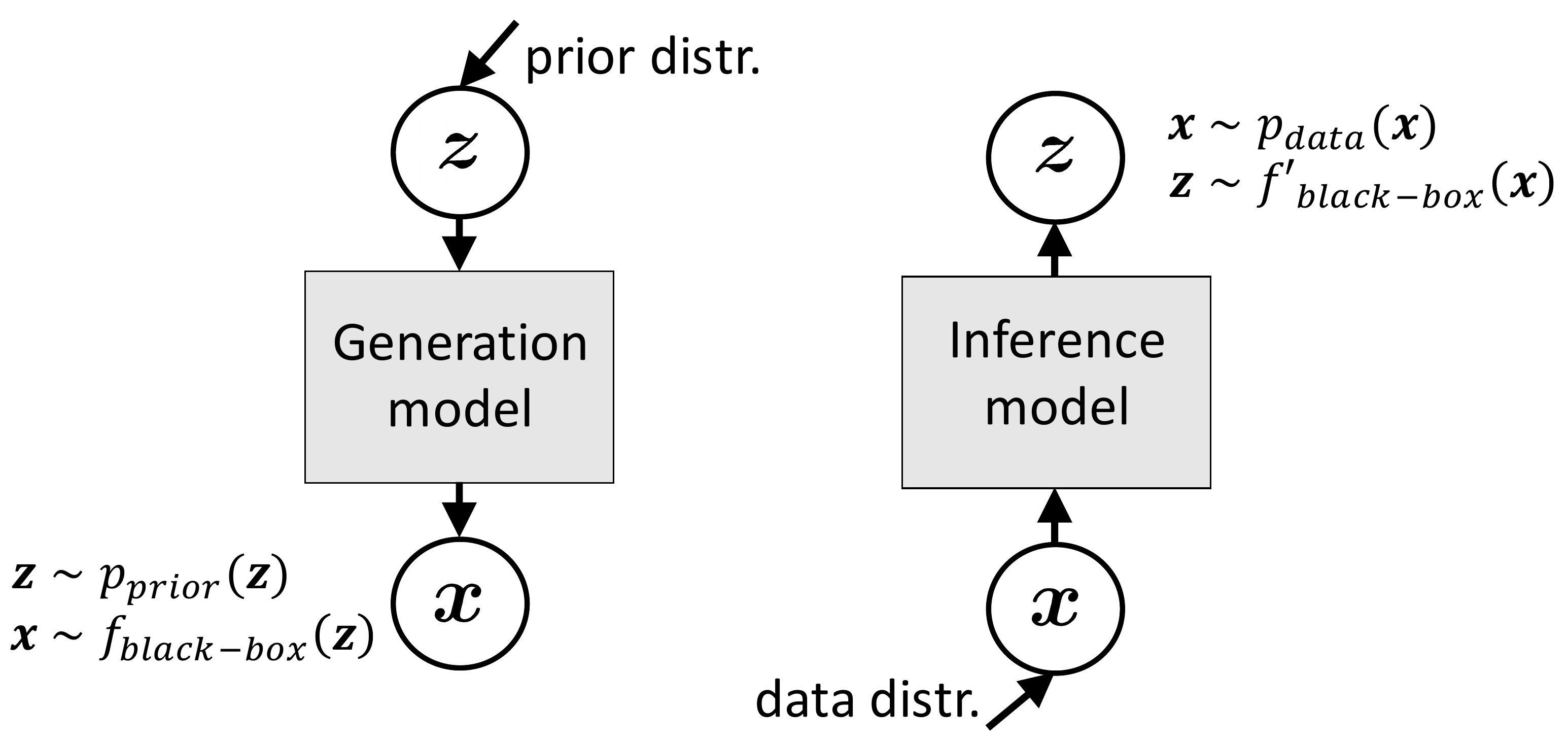}
  \vspace{-5pt}
  \caption{Symmetric view of generation and inference. There is little difference of the two processes in terms of formulation: with implicit distribution modeling, both processes only need to perform simulation through black-box neural transformations between the latent and visible spaces.} 
  \label{fig:symmetric}
\end{figure}

Our new interpretations of GANs and VAEs have revealed strong connections between them. One of the key ideas in our formulation is to interpret sample generation in GANs as performing posterior inference. This section provides a more general discussion of this point.

Traditional modeling approaches usually distinguish between latent and visible variables clearly and treat them in very different ways. One of the key thoughts in our formulation is that it is not necessary to make clear boundary between the two types of variables (and between generation and inference), but instead, treating them as a symmetric pair helps with modeling and understanding (Figure~\ref{fig:symmetric}). For instance, we treat the generation space $\bm{x}$ in GANs as latent, which immediately reveals the connection between GANs and adversarial domain adaptation, and provides a variational inference interpretation of the generation. A second example is the classic wake-sleep algorithm, where the wake phase reconstructs visibles conditioned on latents, while the sleep phase reconstructs latents conditioned on visibles (i.e., generated samples). Hence, visible and latent variables are treated in a completely symmetric manner. 

The (empirical) data distributions over visible variables are usually implicit, i.e., easy to sample from but intractable for evaluating likelihood. In contrast, the prior distributions over latent variables are usually defined as explicit distributions, amenable to likelihood evaluation. Fortunately, the adversarial approach in GANs, and other techniques such as density ratio estimation~\citep{mohamed2016learning} and approximate Bayesian computation~\citep{beaumont2002approximate}, have provided useful tools to bridge the gap. For instance, implicit generative models such as GANs require only simulation of the generative process without explicit likelihood evaluation, hence the prior distributions over latent variables are used in the same way as the empirical data distributions, namely, simulating samples. For explicit likelihood-based models, adversarial autoencoder (AAE) leverages the adversarial approach to allow implicit prior distributions over latent space. 
Likewise, the implicit variational inference methods~(section~\ref{sec:implicit-inference}) either do not require explicit distributions as priors. In these methods, adversarial loss is used to replace intractable KL divergence loss between the variational distributions and the priors. In sum, with the new tools like the adversarial loss, prior distributions over latent variables can be (used as) implicit distributions precisely the same as the empirical data distribution.

The second difference between the visible and latent variables involves the complexity of the two spaces. Visible space is usually complex while latent space tends (or is designed) to be simpler. The complexity difference guides us to choose appropriate tools (e.g., adversarial loss v.s. maximum likelihood loss, etc) to minimize the distance between distributions to learn and their targets. 
For instance, VAEs and adversarial autoencoder both regularize the model by minimizing the distance between the variational posterior and the prior, though VAEs choose KL divergence loss while AAE selects adversarial loss. 

We can further extend the symmetric treatment of visible/latent $\bm{x}$/$\bm{z}$ pair to data/label $\bm{x}$/$\bm{t}$ pair, leading to a unified view of the generative and discriminative paradigms for unsupervised and semi-supervised learning. Specifically, conditional generative models create (data, label) pairs by generating data $\bm{x}$ given label $\bm{t}$. These pairs can be used for classifier training~\citep{hu2017controllable,odena2017conditional}. In parallel, discriminative approaches such as knowledge distillation~\citep{hinton2015distilling,hu2016harnessing,hu2016deep} create (data, label) pairs by generating label $\bm{t}$ given data $\bm{x}$. With the symmetric view of the $\bm{x}$- and $\bm{t}$-spaces, and neural network-based black-box mappings between spaces, we can see the two approaches are essentially the same.

{\small
\bibliography{refs}
\bibliographystyle{abbrvnat}
}

\newpage

\appendix

\section{Proof of Lemma 1}\label{supp:sec:lemma-1}
\begin{proof}
\begin{equation}
\small
\begin{split}
& \E_{p_{\theta}(\bm{x}|y)p(y)}\left[ \log q^{r}(y|\bm{x}) \right] = \\
& - \E_{p(y)}\left[ \KL\left( p_{\theta}(\bm{x}|y) \| q^{r}(\bm{x}|y)\right) - \KL(p_{\theta}(\bm{x}|y)\|p_{\theta_0}(\bm{x})) \right], \\
\end{split}
\label{supp:eq:thm-1-proof-ext}
\end{equation}
where
\begin{equation}
\small
\begin{split}
&\E_{p(y)}\left[ \KL(p_{\theta}(\bm{x}|y)\|p_{\theta_0}(\bm{x})) \right] \\
&= p(y=0)\cdot \KL\left(p_{\theta}(\bm{x}|y=0) \| \frac{p_{\theta_0}(\bm{x}|y=0)+p_{\theta_0}(\bm{x}|y=1)}{2}\right) \\ 
&\quad +  p(y=1)\cdot \KL\left(p_{\theta}(\bm{x}|y=1) \| \frac{p_{\theta_0}(\bm{x}|y=0)+p_{\theta_0}(\bm{x}|y=1)}{2} \right).
\end{split}
\label{supp:eq:thm-1-jsd-1}
\end{equation}
Note that $p_\theta(\bm{x}|y=0)=p_{g_\theta}(\bm{x})$, and $p_\theta(\bm{x}|y=1)=p_{data}(\bm{x})$. Let $p_{M_\theta}=\frac{p_{g_\theta}+p_{data}}{2}$. Eq.\eqref{supp:eq:thm-1-jsd-1} can be simplified as:
\begin{equation}
\small
\begin{split}
&\E_{p(y)}\left[ \KL(p_{\theta}(\bm{x}|y)\|p_{\theta_0}(\bm{x})) \right] = \frac{1}{2} \KL\left( p_{g_\theta}\|p_{M_{\theta_0}} \right) + \frac{1}{2} \KL\left( p_{data}\|p_{M_{\theta_0}} \right).
\end{split}
\label{supp:eq:thm-1-jsd-2}
\end{equation}
On the other hand,
\begin{equation}
\small
\begin{split}
\JSD(p_{g_\theta}\|p_{data}) &= \frac{1}{2} \E_{p_{g_\theta}}\left[ \log \frac{p_{g_\theta}}{p_{M_\theta}} \right] + \frac{1}{2} \E_{p_{data}}\left[ \log \frac{p_{data}}{p_{M_\theta}} \right] \\
&= \frac{1}{2} \E_{p_{g_\theta}}\left[\log \frac{p_{g_\theta}}{p_{M_{\theta_0}}} \right] + \frac{1}{2} \E_{p_{g_\theta}}\left[ \log \frac{p_{M_{\theta_0}}}{p_{M_{\theta}}} \right] \\ 
&\quad + \frac{1}{2} \E_{p_{data}}\left[\log \frac{p_{data}}{p_{M_{\theta_0}}} \right] + \frac{1}{2} \E_{p_{data}}\left[ \log \frac{p_{M_{\theta_0}}}{p_{M_{\theta}}} \right] \\
&= \frac{1}{2} \E_{p_{g_\theta}}\left[\log \frac{p_{g_\theta}}{p_{M_{\theta_0}}} \right] + \frac{1}{2} \E_{p_{data}}\left[\log \frac{p_{data}}{p_{M_{\theta_0}}} \right] + \E_{p_{M_\theta}}\left[\log \frac{p_{M_{\theta_0}}}{p_{M_{\theta}}} \right] \\
&= \frac{1}{2} \KL\left( p_{g_\theta}\|p_{M_{\theta_0}} \right) + \frac{1}{2} \KL\left( p_{data}\|p_{M_{\theta_0}}\right) - \KL\left( p_{M_\theta}\|p_{M_{\theta_0}} \right).
\end{split}
\label{supp:eq:thm-1-jsd-expand}
\end{equation}
Note that
\begin{equation}
\small
\begin{split}
\nabla_{\theta} \KL\left( p_{M_\theta}\|p_{M_{\theta_0}} \right) \left|_{\theta=\theta_0}\right. = 0.
\end{split}
\end{equation}
Taking derivatives of Eq.\eqref{supp:eq:thm-1-jsd-2} w.r.t $\bm{\theta}$ at $\bm{\theta}_0$ we get
\begin{equation}
\small
\begin{split}
&\nabla_{\theta} \E_{p(y)}\left[ \KL(p_{\theta}(\bm{x}|y)\|p_{\theta_0}(\bm{x})) \right] \left|_{\theta=\theta_0}\right. \\
&= \nabla_{\theta} \left( \frac{1}{2} \KL\left( p_{g_\theta}\|p_{M_{\theta_0}} \right)\left|_{\theta=\theta_0}\right. +  \frac{1}{2} \KL\left( p_{data}\|p_{M_{\theta_0}}\right) \right) \left|_{\theta=\theta_0}\right. \\
&=  \nabla_{\theta} \JSD(p_{g_\theta}\|p_{data}) \left|_{\theta=\theta_0}\right..
\end{split}
\label{supp:eq:thm-1-dev-2nd}
\end{equation}

Taking derivatives of the both sides of Eq.\eqref{supp:eq:thm-1-proof-ext} at w.r.t $\bm{\theta}$ at $\bm{\theta}_0$ and plugging the last equation of Eq.\eqref{supp:eq:thm-1-dev-2nd}, we obtain the desired results.
\end{proof}

\section{Proof of JSD Upper Bound in Lemma~1}\label{supp:sec:jsd-bound}
We show that, in Lemma.1 (Eq.\ref{eq:thm-1-result}), the JSD term is upper bounded by the KL term, i.e.,
\begin{equation}
\begin{split}
\JSD(p_\theta(\bm{x}|y=0)\|p_\theta(\bm{x}|y=1)) \leq \E_{p(y)}\left[ \KL(p_\theta(\bm{x}|y) \| q^r(\bm{x}|y)) \right].
\end{split}
\label{supp:eq:jsd-bound}
\end{equation}
\begin{proof}
From Eq.\eqref{supp:eq:thm-1-proof-ext}, we have
\begin{equation}
\begin{split}
\E_{p(y)}\left[ \KL(p_{\theta}(\bm{x}|y)\|p_{\theta_0}(\bm{x})) \right] \leq \E_{p(y)}\left[ \KL\left( p_{\theta}(\bm{x}|y) \| q^{r}(\bm{x}|y)\right) \right].
\end{split}
\label{supp:eq:jsd-bound-1}
\end{equation}
From Eq.\eqref{supp:eq:thm-1-jsd-2} and Eq.\eqref{supp:eq:thm-1-jsd-expand}, we have
\begin{equation}
\begin{split}
\JSD(p_\theta(\bm{x}|y=0)\|p_\theta(\bm{x}|y=1)) \leq \E_{p(y)}\left[ \KL(p_{\theta}(\bm{x}|y)\|p_{\theta_0}(\bm{x})) \right].
\end{split}
\label{supp:eq:jsd-bound-2}
\end{equation}
Eq.\eqref{supp:eq:jsd-bound-1} and Eq.\eqref{supp:eq:jsd-bound-2} lead to Eq.\eqref{supp:eq:jsd-bound}.
\end{proof}

\begin{figure} 
  \centering 
  \includegraphics[width=0.7\linewidth]{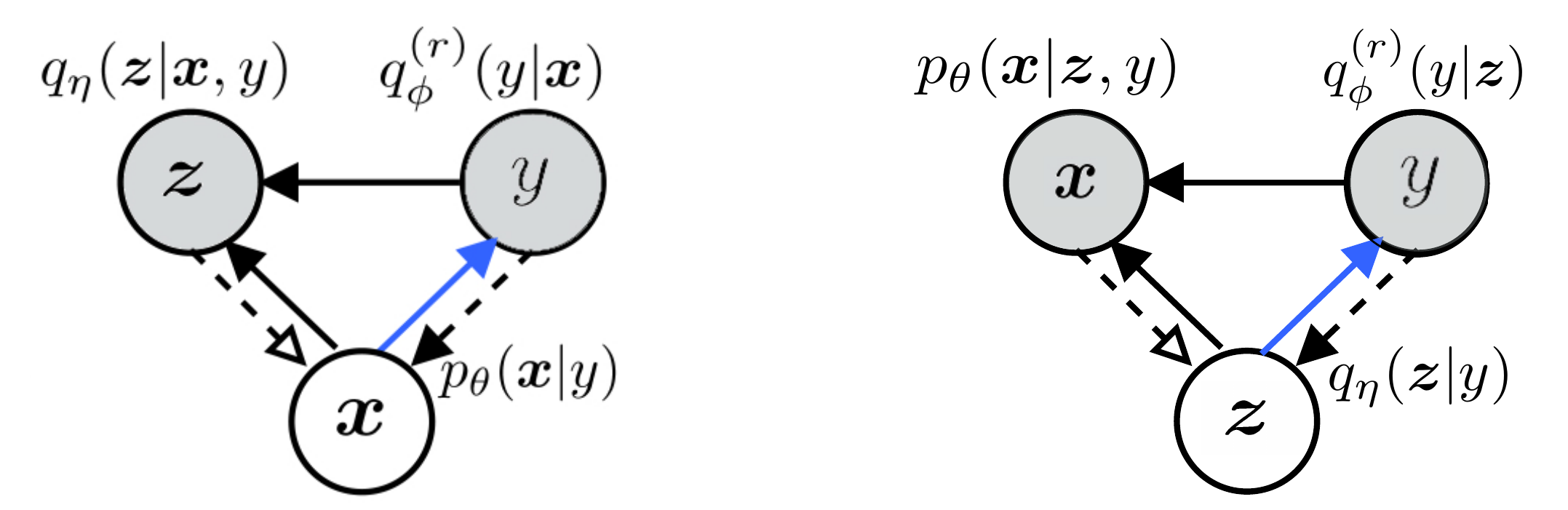}
  %\vspace{-10pt}
  \caption{{\bf Left:} Graphical model of InfoGAN. {\bf Right:}  Graphical model of Adversarial Autoencoder (AAE), which is obtained by swapping data $\bm{x}$ and code $\bm{z}$ in InfoGAN.} 
  \label{supp:fig:gm_infogan_aa}
\end{figure}

\section{Schematic Graphical Models and AAE/PM/CycleGAN}\label{supp:sec:aae}
{\bf Adversarial Autoencoder (AAE)}~\citep{makhzani2015adversarial} can be obtained by swapping code variable $\bm{z}$ and data variable $\bm{x}$ of InfoGAN in the graphical model, as shown in Figure~\ref{supp:fig:gm_infogan_aa}. To see this, we directly write down the objectives represented by the graphical model in the right panel, and show they are precisely the original AAE objectives proposed in ~\citep{makhzani2015adversarial}. We present detailed derivations, which also serve as an example for how one can translate a graphical model representation to the mathematical formulations. Readers can do similarly on the schematic graphical models of GANs, InfoGANs, VAEs, and many other relevant variants and write down the respective objectives conveniently.

We stick to the notational convention in the paper that parameter $\bm{\theta}$ is associated with the distribution over $\bm{x}$, parameter $\bm{\eta}$ with the distribution over $\bm{z}$, and parameter $\bm{\phi}$ with the distribution over $y$. Besides, we use $p$ to denote the distributions over $\bm{x}$, and $q$ the distributions over $\bm{z}$ and $y$. 

From the graphical model, the inference process (dashed-line arrows) involves implicit distribution $q_{\eta}(\bm{z}|y)$ (where $\bm{x}$ is encapsulated). As in the formulations of GANs (Eq.4 in the paper) and VAEs (Eq.13 in the paper), $y=1$ indicates the real distribution we want to approximate and $y=0$ indicates the approximate distribution with parameters to learn. So we have
\begin{equation}
\small
\begin{split}
q_\eta(\bm{z}|y) = 
\begin{cases}
q_\eta(\bm{z}|y=0) & y=0 \\
q(\bm{z})  &  y=1,
\end{cases}
\end{split}
\label{supp:eq:aae-pz_y}
\end{equation}
where, as $\bm{z}$ is the hidden code, $q(\bm{z})$ is the prior distribution over $\bm{z}$\footnote{See section~6 of the paper for the detailed discussion on prior distributions of hidden variables and empirical distribution of visible variables}, and the space of $\bm{x}$ is degenerated. Here $q_\eta(\bm{z}|y=0)$ is the implicit distribution such that
\begin{equation}
\small
\begin{split}
\bm{z} \sim q_\eta(\bm{z}|y=0)\quad \Longleftrightarrow\quad \bm{z} = E_{\eta}(\bm{x}),~~ \bm{x}\sim p_{data}(\bm{x}),
\end{split}
\label{supp:eq:aae-pz_y_0}
\end{equation}
where $E_{\eta}(\bm{x})$ is a deterministic transformation parameterized with $\bm{\eta}$ that maps data $\bm{x}$ to code $\bm{z}$. Note that as $\bm{x}$ is a visible variable, the pre-fixed distribution of $\bm{x}$ is the empirical data distribution. 

On the other hand, the generative process (solid-line arrows) involves $p_\theta(\bm{x}|\bm{z},y)q^{(r)}_\phi(y|\bm{z})$ (here $q^{(r)}$ means we will swap between $q^{r}$ and $q$). As the space of $\bm{x}$ is degenerated given $y=1$, thus $p_\theta(\bm{x}|\bm{z},y)$ is fixed without parameters to learn, and $\bm{\theta}$ is only associated to $y=0$.

With the above components, we maximize the log likelihood of the generative distributions \\
$\log p_\theta(\bm{x}|\bm{z},y)q^{(r)}_\phi(y|\bm{z})$ conditioning on the variable $\bm{z}$ inferred by $q_{\eta}(\bm{z}|y)$. Adding the prior distributions, the objectives are then written as 
\begin{equation}
\small
\begin{split}
&\max\nolimits_{\bm{\phi}} \loss_{\phi} = \E_{q_\eta(\bm{z}|y)p(y)}\left[ \log p_{\theta}(\bm{x}|\bm{z}, y) q_\phi(y|\bm{z})  \right] \\
&\max\nolimits_{\bm{\theta},\bm{\eta}} \loss_{\theta,\eta} = \E_{q_\eta(\bm{z}|y)p(y)}\left[ \log p_{\theta}(\bm{x}|\bm{z}, y) q^{r}_\phi(y|\bm{z})  \right].
\end{split}
\label{supp:eq:aae-obj}
\end{equation}
Again, the only difference between the objectives of $\bm{\phi}$ and $\{\bm{\theta},\bm{\eta}\}$ is swapping between $q_\phi(y|\bm{z})$ and its reverse $q^{r}_\phi(y|\bm{z})$.

To make it clearer that Eq.\eqref{supp:eq:aae-obj} is indeed the original AAE proposed in~\citep{makhzani2015adversarial}, we transform $\loss_{\phi}$ as
\begin{equation}
\small
\begin{split}
\max\nolimits_{\bm{\phi}} \loss_{\phi} &= \E_{q_\eta(\bm{z}|y)p(y)}\left[ \log q_\phi(y|\bm{z})  \right] \\
&=\frac{1}{2}\E_{q_\eta(\bm{z}|y=0)}\left[ \log q_\phi(y=0|\bm{z})  \right] + \frac{1}{2}\E_{q_\eta(\bm{z}|y=1)}\left[ \log q_\phi(y=1|\bm{z})  \right] \\
&= \frac{1}{2}\E_{\bm{z}=E_\eta(\bm{x}), \bm{x}\sim p_{data}(\bm{x})} \left[ \log q_\phi(y=0|\bm{z}) \right] + \frac{1}{2}\E_{\bm{z}\sim q(\bm{z})}\left[ \log q_\phi(y=1|\bm{z})  \right].
\end{split}
\label{supp:eq:aae-obj-phi}
\end{equation}
That is, the discriminator with parameters $\bm{\phi}$ is trained to maximize the accuracy of distinguishing the hidden code either sampled from the true prior $p(\bm{z})$ or inferred from observed data example $\bm{x}$. The objective $\loss_{\theta,\eta}$ optimizes $\bm{\theta}$ and $\bm{\eta}$ to minimize the reconstruction loss of observed data $\bm{x}$ and at the same time to generate code $\bm{z}$ that fools the discriminator. We thus get the conventional view of the AAE model.

{\bf Predictability Minimization (PM)}~\citep{schmidhuber1992learning} is the early form of adversarial approach which aims at learning code $\bm{z}$ from data such that each unit of the code is hard to predict by the accompanying code predictor based on remaining code units. AAE closely resembles PM by seeing the discriminator as a special form of the code predictors.

%{\bf CycleGAN}~\citep{zhu2017unpaired} is the model that learns to translate examples of one domain (e.g., images of horse) to another domain (e.g., images of zebra) and vice versa based on unpaired data. Let $\bm{x}$ and $\bm{z}$ be the variables of the two domains, then the objectives of AAE (Eq.\ref{supp:eq:aae-obj}) is precisely the objectives that train the model to translate $\bm{x}$ into $\bm{z}$. The reversed translation is trained with the objectives of InfoGAN (Eq.9 in the paper), the symmetric counterpart of AAE. 

\section{Proof of Lemme 2}\label{supp:sec:lemma-2}
\begin{proof}
For the reconstruction term:
\begin{equation}
\small
\begin{split}
&\E_{p_{\theta_0}(\bm{x})}\left[ \E_{\begin{subarray}{l} q_{\eta}(\bm{z}|\bm{x},y) q_{*}^{r}(y|\bm{x}) \end{subarray}} \left[ \log p_\theta(\bm{x}|\bm{z},y) \right] \right] \\
&= \frac{1}{2}\E_{p_{\theta_0}(\bm{x}|y=1)}\left[ \E_{\begin{subarray}{l} q_{\eta}(\bm{z}|\bm{x},y=0), y=0\sim q_{*}^{r}(y|\bm{x}) \end{subarray}} \left[ \log p_\theta(\bm{x}|\bm{z},y=0) \right] \right] \\
&+ \frac{1}{2}\E_{p_{\theta_0}(\bm{x}|y=0)}\left[ \E_{\begin{subarray}{l} q_{\eta}(\bm{z}|\bm{x},y=1), y=1\sim q_{*}^{r}(y|\bm{x}) \end{subarray}} \left[ \log p_\theta(\bm{x}|\bm{z},y=1) \right] \right] \\
&= \frac{1}{2}\E_{p_{data}(\bm{x})}\left[ \E_{\begin{subarray}{l} \tilde{q}_{\eta}(\bm{z}|\bm{x}) \end{subarray}} \left[ \log \tilde{p}_\theta(\bm{x}|\bm{z}) \right] \right] + const,
\end{split}
\label{supp:eq:lemma-proof-recon}
\end{equation}
where $y=0\sim q^{r}_*(y|\bm{x})$ means $q^{r}_*(y|\bm{x})$ predicts $y=0$ with probability $1$. Note that both $q_{\eta}(\bm{z}|\bm{x},y=1)$ and $p_\theta(\bm{x}|\bm{z},y=1)$ are constant distributions without free parameters to learn; $q_{\eta}(\bm{z}|\bm{x},y=0)=\tilde{q}_{\eta}(\bm{z}|\bm{x})$, and $p_\theta(\bm{x}|\bm{z},y=0)=\tilde{p}_\theta(\bm{x}|\bm{z})$. 

For the KL prior regularization term:
\begin{equation}
\small
\begin{split}
&\E_{p_{\theta_0}(\bm{x})}\left[ \KL(q_\eta(\bm{z}|\bm{x},y) q_{*}^{r}(y|\bm{x}) \| p(\bm{z}|y)p(y)) \right] \\
&= \E_{p_{\theta_0}(\bm{x})}\left[ \int q^r_*(y|\bm{x}) \KL\left( q_\eta(\bm{z}|\bm{x},y) \| p(\bm{z}|y) \right)dy + \KL\left(q^r_*(y|\bm{x})\|p(y)\right) \right] \\
&= \frac{1}{2}\E_{p_{\theta_0}(\bm{x}|y=1)}\left[ \KL\left( q_\eta(\bm{z}|\bm{x},y=0) \| p(\bm{z}|y=0) \right) + const \right] + \frac{1}{2}\E_{p_{\theta_0}(\bm{x}|y=1)}\left[const\right] \\
&= \frac{1}{2}\E_{p_{data}(\bm{x})}\left[ \KL(\tilde{q}_\eta(\bm{z}|\bm{x}) \| \tilde{p}(\bm{z})) \right].
\end{split}
\label{supp:eq:lemma-proof-kl}
\end{equation}
Combining Eq.\eqref{supp:eq:lemma-proof-recon} and Eq.\eqref{supp:eq:lemma-proof-kl} we recover the conventional VAE objective in Eq.(7) in the paper.
\end{proof}

\section{Importance Weighted GANs (IWGAN)}
From Eq.\eqref{eq:thm-1-result} in the paper, we can view GANs as maximizing a lower bound of the ``marginal log-likelihood'' on $y$: 
\begin{equation}
\small
\begin{split}
\log q(y) &= \log \int p_{\theta}(\bm{x}|y)  \frac{q^r(y|\bm{x})p_{\theta_0}(\bm{x})}{p_\theta(\bm{x}|y)} d\bm{x} \\
&\geq \int p_{\theta}(\bm{x}|y) \log \frac{q^r(y|\bm{x})p_{\theta_0}(\bm{x})}{p_\theta(\bm{x}|y)} d\bm{x} \\
&= -\KL(p_{\theta}(\bm{x}|y) \| q^r(\bm{x}|y)) + const.
\end{split}
\label{supp:eq:iwgan-ori}
\end{equation}
We can apply the same importance weighting method as in IWAE~\citep{burda2015importance} to derive a tighter bound.
\begin{equation}
\small
\begin{split}
\log q(y) &= \log \E\left[ \frac{1}{k}\sum_{i=1}^{k} \frac{q^r(y|\bm{x}_i)p_{\theta_0}(\bm{x}_i)}{p_\theta(\bm{x}_i|y)} \right] \\
&\geq \E\left[ \log \frac{1}{k}\sum_{i=1}^{k} \frac{q^r(y|\bm{x}_i)p_{\theta_0}(\bm{x}_i)}{p_\theta(\bm{x}_i|y)} \right] \\
&= \E\left[ \log \frac{1}{k}\sum_{i=1}^{k} w_i \right] \\
&:= \mathcal{L}_k(y)
\end{split}
\label{supp:eq:iwgan}
\end{equation}
where we have denoted $w_i = \frac{q^r(y|\bm{x}_i)p_{\theta_0}(\bm{x}_i)}{p_\theta(\bm{x}_i|y)}$, which is the unnormalized importance weight. We recover the lower bound of Eq.\eqref{supp:eq:iwgan-ori} when setting $k=1$.

To maximize the importance weighted lower bound $\mathcal{L}_k(y)$, we take the derivative w.r.t $\bm{\theta}$ and apply the reparameterization trick on samples $\bm{x}$:
\begin{equation}
\small
\begin{split}
\nabla_\theta \mathcal{L}_k(y) = \nabla_\theta \E_{\bm{x}_1,\dots,\bm{x}_k}\left[ \log \frac{1}{k}\sum_{i=1}^{k} w_i \right] &= \E_{\bm{z}_1,\dots,\bm{z}_k} \left[ \nabla_\theta \log \frac{1}{k}\sum_{i=1}^{k} w(y, \bm{x}(\bm{z}_i,\bm{\theta})) \right] \\
&= \E_{\bm{z}_1,\dots,\bm{z}_k} \left[ \sum_{i=1}^{k} \widetilde{w_i} \nabla_\theta \log  w(y, \bm{x}(\bm{z}_i,\bm{\theta})) \right],
\end{split}
\label{supp:eq:iwgan-grad}
\end{equation}
where $\widetilde{w_i} = w_i / \sum_{i=1}^k w_i$ are the normalized importance weights. We expand the weight at $\bm{\theta}=\bm{\theta}_0$
\begin{equation}
\small
\begin{split}
w_i |_{\theta=\theta_0} = \frac{q^r(y|\bm{x}_i)p_{\theta_0}(\bm{x}_i)}{p_\theta(\bm{x}_i|y)} &= q^r(y|\bm{x}_i) \frac{\frac{1}{2}p_{\theta_0}(\bm{x}_i|y=0) + \frac{1}{2}p_{\theta_0}(\bm{x}_i|y=1)}{p_{\theta_0}(\bm{x}_i|y)} |_{\theta=\theta_0}.  
\end{split}
\label{supp:eq:iwgan-w}
\end{equation}
The ratio of $p_{\theta_0}(\bm{x}_i|y=0)$ and $p_{\theta_0}(\bm{x}_i|y=1)$ is intractable. Using the Bayes' rule and approximating with the discriminator distribution, we have
\begin{equation}
\small
\begin{split}
\frac{p(\bm{x}|y=0)}{p(\bm{x}|y=1)} = \frac{p(y=0|\bm{x})p(y=1)}{p(y=1|\bm{x})p(y=0)} \approx \frac{q(y=0|\bm{x})}{q(y=1|\bm{x})}.
\end{split}
\label{supp:eq:iwgan-w-bayes}
\end{equation}
Plug Eq.\eqref{supp:eq:iwgan-w-bayes} into the above we have
\begin{equation}
\small
\begin{split}
w_i |_{\theta=\theta_0} \approx \frac{q^r(y|\bm{x}_i)}{q(y|\bm{x}_i)}.
\end{split}
\label{supp:eq:iwgan-w-approx}
\end{equation}

In Eq.\eqref{supp:eq:iwgan-grad}, the derivative $\nabla_\theta \log w_i$ is
\begin{equation}
\small
\begin{split}
\nabla_\theta \log  w(y, \bm{x}(\bm{z}_i,\bm{\theta})) = \nabla_\theta \log q^r(y|\bm{x}(\bm{z}_i,\bm{\theta})) +  \nabla_\theta \log \frac{p_{\theta_0}(\bm{x}_i)}{p_\theta(\bm{x}_i|y)}.
\end{split}
\label{supp:eq:iwgan-grad-expand}
\end{equation}
The second term in the RHS of the equation is intractable as it involves evaluating the likelihood of implicit distributions. However, if we take $k=1$, it can be shown that
\begin{equation}
\small
\begin{split}
&- \E_{p(y)p(\bm{z}|y)}\left[ \nabla_\theta \log \frac{p_{\theta_0}(\bm{x}(\bm{z},\bm{\theta}))}{p_\theta(\bm{x}(\bm{z},\bm{\theta})|y)} |_{\bm{\theta}=\bm{\theta}_0}\right] \\
&= - \nabla_\theta \frac{1}{2} \E_{p_\theta(\bm{x}|y=0)}\left[ \frac{p_{\theta_0}(\bm{x})}{p_\theta(\bm{x}|y=0)} \right] + \frac{1}{2} \E_{p_\theta(\bm{x}|y=1)}\left[ \frac{p_{\theta_0}(\bm{x})}{p_\theta(\bm{x}|y=1)} \right] |_{\theta=\theta_0} \\
&= \nabla_\theta \JSD(p_{g_\theta}(\bm{x})\|p_{data}(\bm{x})) |_{\theta=\theta_0},
\end{split}
\label{supp:eq:iwgan-grad-2nd-jsd}
\end{equation}
where the last equation is based on Eq.\eqref{supp:eq:thm-1-jsd-expand}. That is, the second term in the RHS of Eq.\eqref{supp:eq:iwgan-grad-expand} is (when $k=1$) indeed the gradient of the JSD, which is subtracted away in the standard GANs as shown in Eq.\eqref{eq:thm-1-result} in the paper. We thus follow the standard GANs and also remove the second term even when $k>1$.
Therefore, the resulting update rule for the generator parameter $\bm{\theta}$ is
\begin{equation}
\small
\begin{split}
\nabla_\theta \mathcal{L}_k(y) = \E_{\bm{z}_1,\dots,\bm{z}_k \sim p(\bm{z}|y)} \left[ \sum\nolimits_{i=1}^{k} \widetilde{w_i} \nabla_\theta \log  q^r_{\phi_0}(y|\bm{x}(\bm{z}_i,\bm{\theta})) \right].
\end{split}
\label{eq:iwgan-update}
\end{equation}

\section{Adversary Activated VAEs (AAVAE)}
In our formulation, VAEs include a degenerated adversarial discriminator which blocks out generated samples from contributing to model learning. We enable adaptive incorporation of fake samples by activating the adversarial mechanism. Again, derivations are straightforward by making symbolic analog to GANs. 

We replace the perfect discriminator $q_{*}(y|\bm{x})$ in vanilla VAEs with the discriminator network $q_{\phi}(y|\bm{x})$ parameterized with $\bm{\phi}$ as in GANs, resulting in an adapted objective of Eq.\eqref{eq:vae-new} in the paper:
\begin{equation}
\small
\begin{split}
\max\nolimits_{\bm{\theta},\bm{\eta}} \mathcal{L}^{\text{aavae}}_{\theta,\eta} &=  \E_{p_{\theta_0}(\bm{x})}\left[ \E_{\begin{subarray}{l} q_{\eta}(\bm{z}|\bm{x},y) q_{\phi}^{r}(y|\bm{x}) \end{subarray}} \left[ \log p_\theta(\bm{x}|\bm{z},y) \right] - \KL(q_\eta(\bm{z}|\bm{x},y) q_{\phi}^{r}(y|\bm{x}) \| p(\bm{z}|y)p(y)) \right].
\end{split}
\label{supp:eq:aavae-gen}
\end{equation}
The form of Eq.\eqref{supp:eq:aavae-gen} is precisely symmetric to the objective of InfoGAN in Eq.\eqref{eq:infogan-obj} with the additional KL prior regularization. Before analyzing the effect of adding the learnable discriminator, we first look at how the discriminator is learned. In analog to GANs in Eq.\eqref{eq:ada-obj} and InfoGANs in Eq.\eqref{eq:infogan-obj}, the objective of optimizing $\bm{\phi}$ is obtained by simply replacing the inverted distribution $q_{\phi}^{r}(y|\bm{x})$ with $q_{\phi}(y|\bm{x})$:
\begin{equation}
\small
\begin{split}
\max\nolimits_{\bm{\phi}} \mathcal{L}^{\text{aavae}}_{\phi} &=  \E_{p_{\theta_0}(\bm{x})}\left[ \E_{\begin{subarray}{l} q_{\eta}(\bm{z}|\bm{x},y) q_{\phi}(y|\bm{x}) \end{subarray}} \left[ \log p_\theta(\bm{x}|\bm{z},y) \right] - \KL(q_\eta(\bm{z}|\bm{x},y) q_{\phi}(y|\bm{x}) \| p(\bm{z}|y)p(y)) \right].
\end{split}
\label{supp:eq:aavae-dis}
\end{equation}
Intuitively, the discriminator is trained to distinguish between real and fake instances by predicting appropriate $y$ that selects the components of $q_\eta(\bm{z}|\bm{x},y)$ and $p_\theta(\bm{x}|\bm{z},y)$ to best reconstruct $\bm{x}$.
The difficulty of Eq.\eqref{supp:eq:aavae-dis} is that $p_{\theta}(\bm{x}|\bm{z},y=1)=p_{data}(\bm{x})$ is an implicit distribution which is intractable for likelihood evaluation. We thus use the alternative objective as in GANs to train a binary classifier:
\begin{equation}
\small
\begin{split}
\max\nolimits_{\bm{\phi}} \mathcal{L}^{\text{aavae}}_{\phi} &=  \E_{p_{\theta}(\bm{x}|\bm{z},y)p(\bm{z}|y)p(y)} \left[ \log q_\phi(y|\bm{x}) \right].
\end{split}
\label{supp:eq:aavae-dis-gan}
\end{equation}

\section{Experiments}

\subsection{Importance Weighted GANs}
We extend both vanilla GANs and class-conditional GANs (CGAN) with the importance weighting method.
The base GAN model is implemented with the DCGAN architecture and hyperparameter setting~\citep{radford2015unsupervised}. We do not tune the hyperparameters for the importance weighted extensions. We use MNIST, SVHN, and CIFAR10 for evaluation. For vanilla GANs and its IW extension, we measure inception scores~\citep{salimans2016improved} on the generated samples. We train deep residual networks provided in the tensorflow library as evaluation networks, which achieve inception scores of $9.09$, $6.55$, and $8.77$ on the test sets of MNIST, SVHN, and CIFAR10, respectively. For conditional GANs we evaluate the accuracy of conditional generation~\citep{hu2017controllable}. That is, we generate samples given class labels, and then use the pre-trained classifier to predict class labels of the generated samples. The accuracy is calculated as the percentage of the predictions that match the conditional labels. The evaluation networks achieve accuracy of $0.990$ and $0.902$ on the test sets of MNIST and SVHN, respectively.

%Table~\ref{tab:exp}, left panel, shows the inception scores of GANs and IW-GAN, and the middle panel gives the classification accuracy of the conditional GANs and its importance weighted extension IW-CGAN. We report the averaged results $\pm$ one standard deviation over 5 runs. We see that the importance weighting strategy gives consistent improvements over the base models.

\subsection{Adversary Activated VAEs}
We apply the adversary activating method on vanilla VAEs, class-conditional VAEs (CVAE), and semi-supervised VAEs (SVAE)~\citep{kingma2014semi}. We evaluate on the MNIST data. The generator networks have the same architecture as the generators in GANs in the above experiments, with sigmoid activation functions on the last layer to compute the means of Bernoulli distributions over pixels. The inference networks, discriminators, and the classifier in SVAE share the same architecture as the discriminators in the GAN experiments. 

We evaluate the lower bound value on the test set, with varying number of real training examples. For each minibatch of real examples we generate equal number of fake samples for training. In the experiments we found it is generally helpful to smooth the discriminator distributions by setting the temperature of the output sigmoid function larger than 1. This basically encourages the use of fake data for learning. We select the best temperature from $\{1, 1.5, 3, 5\}$ through cross-validation. We do not tune other hyperparameters for the adversary activated extensions. 

Table~\ref{supp:tab:svae} reports the full results of SVAE and AA-SVAE, with the average classification accuracy and standard deviations over 5 runs.
\begin{table}[!h]
\centering
\small
\begin{tabular}{r l l}
\cmidrule[\heavyrulewidth](lr){1-3}
& 1\% & 10\% \\  \cmidrule(lr){1-3}
SVAE &  0.9412$\pm$.0039 & 0.9768$\pm$.0009 \\
AASVAE & {\bf 0.9425$\pm$.0045} &  {\bf 0.9797$\pm$.0010} \\
\cmidrule[\heavyrulewidth](lr){1-3}
\end{tabular}
\caption{Classification accuracy of semi-supervised VAEs and the adversary activated extension on the MNIST test set, with varying size of real labeled training examples.}
\label{supp:tab:svae}
\end{table}

\end{document}